\documentclass[lettersize,journal]{IEEEtran}
\usepackage{amsmath,amsfonts}
\usepackage{algorithmic}
\usepackage{algorithm}
\usepackage{array}
\usepackage[caption=false,font=normalsize,labelfont=sf,textfont=sf]{subfig}
\usepackage{textcomp}
\usepackage{stfloats}
\usepackage{url}
\usepackage{verbatim}
\usepackage{graphicx}
\usepackage{cite}

\usepackage[numbers,sort&compress]{natbib}
\usepackage{bm}
\usepackage{multirow}
\usepackage{amsmath}
\usepackage{amssymb}
\usepackage{pifont}

\newsavebox\CBox

\begin{document}

\title{MM-NeRF: Multimodal-Guided 3D Multi-Style Transfer of Neural Radiance Field}

\author{Zijiang Yang, Zhongwei Qiu, Chang Xu, and Dongmei Fu
\thanks{Zijiang Yang and Dongmei Fu are with the University of Science and Technology Beijing, Beijing 10083, China (e-mail: zijiangyang@xs.ustb.edu.cn; fdm\_ustb@ustb.edu.cn).}
\thanks{Zhongwei Qiu is with the Alibaba DAMO Academy, Hangzhou 311121, China. He is also with the College of Computer Science and Technology, Zhejiang University. (Email: qiuzhongwei.qzw@alibaba-inc.com)}
\thanks{Chang Xu is with the University of Sydney, Darlington, NSW 2008, Australia (e-mail: c.xu@sydney.edu.au).}
\thanks{Corresponding author: Dongmei Fu.}
\thanks{© 2024 IEEE.  Personal use of this material is permitted.  Permission from IEEE must be obtained for all other uses, in any current or future media, including reprinting/republishing this material for advertising or promotional purposes, creating new collective works, for resale or redistribution to servers or lists, or reuse of any copyrighted component of this work in other works.}
}


\markboth{IEEE TRANSACTIONS ON VISUALIZATION AND COMPUTER GRAPHICS,~Vol.~1, No.~1, January~2024}%
{Zijiang Yang, Zhongwei Qiu \MakeLowercase{\textit{et al.}}: MM-NeRF: Multimodal-Guided 3D Multi-Style Transfer of Neural Radiance Field}

\maketitle

\begin{abstract}
3D style transfer aims to generate stylized views of 3D scenes with specified styles, which requires high-quality generating and keeping multi-view consistency.
Existing methods still suffer the challenges of high-quality stylization with texture details and stylization with multimodal guidance.
In this paper, we reveal that the common training method of stylization with NeRF, which generates stylized multi-view supervision by 2D style transfer models, causes the same object in supervision to show various states (color tone, details, etc.) in different views, leading NeRF to tend to smooth the texture details, further resulting in low-quality rendering for 3D multi-style transfer. 
To tackle these problems, we propose a novel Multimodal-guided 3D Multi-style transfer of NeRF, termed MM-NeRF.
First, MM-NeRF projects multimodal guidance into a unified space to keep the multimodal styles consistency and extracts multimodal features to guide the 3D stylization. 
Second, a novel multi-head learning scheme is proposed to relieve the difficulty of learning multi-style transfer, and a multi-view style consistent loss is proposed to track the inconsistency of multi-view supervision data.
Finally, a novel incremental learning mechanism is proposed to generalize MM-NeRF to any new style with small costs.
Extensive experiments on several real-world datasets show that MM-NeRF achieves high-quality 3D multi-style stylization with multimodal guidance, and keeps multi-view consistency and style consistency between multimodal guidance.

\end{abstract}


\section{Introduction}
\IEEEPARstart{R}{ecently}, Neural Radiance Field~\cite{mildenhall2021nerf} (NeRF) has been widely used in 3D representation since it can effectively model complex real-world 3D scenes. Thus, the NeRF-based downstream tasks, such as relighting~\cite{zhang2021nerfactor, srinivasan2021nerv}, surface reconstruction~\cite{wang2021neus}, 3D object generation~\cite{li20223ddesigner}, and 3D style transfer~\cite{zhang2022arf, liu2023stylerf, huang2022stylizednerf} have been rapidly developed.
Besides, with the rapid development of technologies such as Virtual Reality (VR) and Augmented Reality (AR), there has been a diversification in the presentation of 3D content. Among them, 3D style transfer has received widespread attention, which can provide users with a novel artistic experience.
Given reference styles and multi-view images of a 3D scene, 3D style transfer aims to synthesize the novel views of the 3D scene that match the strokes of these styles.

\begin{figure}[t]
    \centering
    \includegraphics[width=0.99\linewidth]{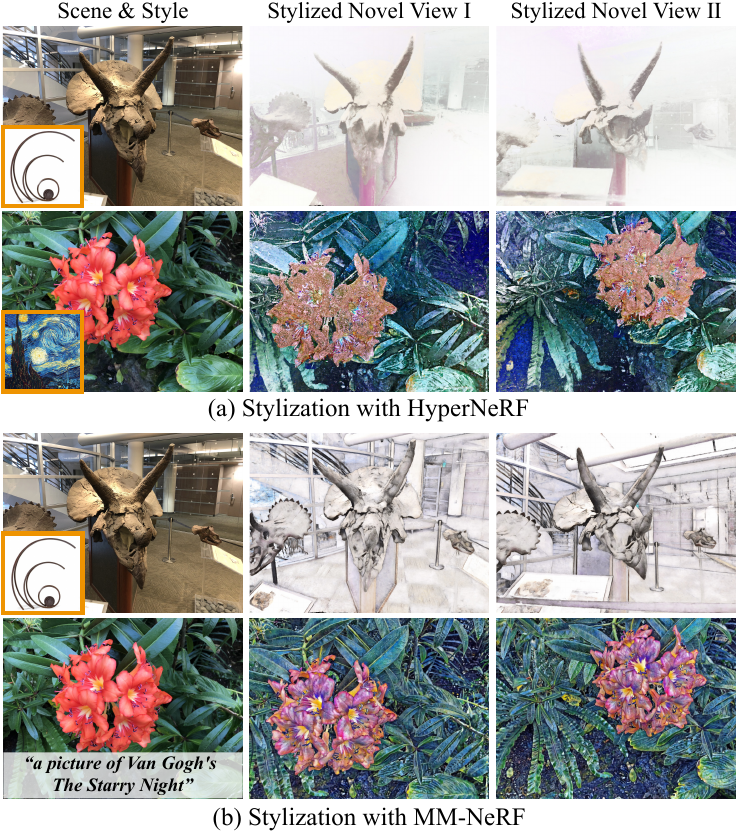}
    \caption{(a) HyperNeRF \cite{chiang2022hyper} generates blurry results and lacks structural details driven by single-modality style (image), (b) MM-NeRF generates higher-quality stylized novel views with multimodal guidance (image and text).
    }
    \label{fig:first_fig}
\end{figure}

Existing methods \cite{chen2022upst, zhang2022arf, nguyen2022snerf, chiang2022hyper, huang2022stylizednerf, liu2023stylerf} have made great progress in 3D style transfer, such as photo-realistic style transfer \cite{chen2022upst}, arbitrary style transfer \cite{liu2023stylerf, chiang2022hyper}, and artistic style transfer \cite{zhang2022arf, nguyen2022snerf, huang2022stylizednerf}.
However, as shown in Figure \ref{fig:first_fig} (a), these methods only focus on the stylization guided by single-modality styles and still suffer the challenges of high-quality stylization. 
This paper aims to achieve multi-style stylization guided by multimodal styles with high-quality details as Figure \ref{fig:first_fig} (b).

\begin{figure}[t]
    \centering
    \includegraphics[width=0.99\linewidth]{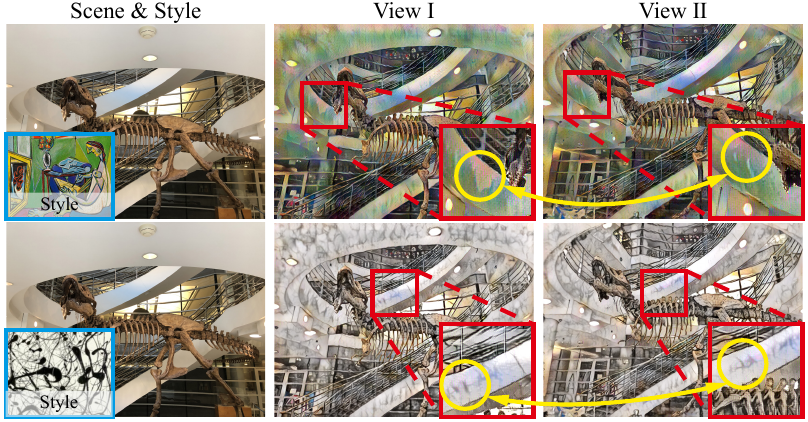}
    \caption{
    Stylized training supervisions generated by 2D style transfer methods. As shown in the zoom-in patch, the local details in different views are stylized inconsistently. In addition, there is a great gap between the two styles, with different tones and strokes.
    }
    \label{fig:motivation}
\end{figure}

\begin{figure}[t]
    \centering
    \includegraphics[width=0.99\linewidth]{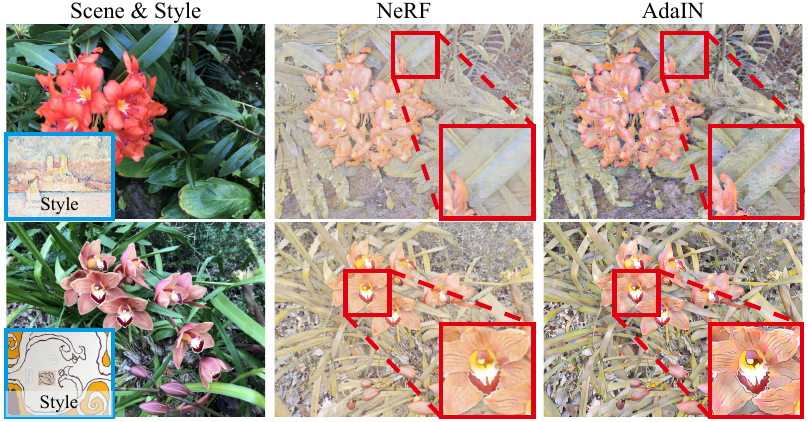}
    \caption{
    Comparison between common NeRF and AdaIN~\cite{huang2017arbitrary}.
    The inconsistency of multi-view supervision makes NeRF generate stylized novel views with blurred details.
    }
    \label{fig:motivation_2}
\end{figure}

Current 3D style transfer methods~\cite{nguyen2022snerf,chen2022upst,huang2022stylizednerf} usually generate stylized multi-view images by 2D stylization models~\cite{huang2017arbitrary,svoboda2020two} and these images are used as the supervisions of 3D style transfer.
However, these methods show blurry details and cannot match the reference styles well.
Two key problems cause this situation.
The first one is inconstant multi-view supervision. As shown in Figure \ref{fig:motivation}, the generated supervisions between different views are inconsistent since each view is stylized independently, which makes NeRF tend to smooth details to reduce total loss, further leading to low-quality stylization. 
The second one is the domain gap among multiple styles. The two styles in Figure \ref{fig:motivation} have extremely different tones and details, which makes it more difficult for neural networks to learn style representations and remains the tendency for style aliasing in multi-style learning.
The two problems bring a challenge for 3D multi-style transfer.
As shown in Figure \ref{fig:motivation_2}, NeRF generates results with blurred details.

In addition, existing methods~\cite{zhang2022arf,chen2022upst,huang2022stylizednerf, liu2023stylerf} only adopt single-modality styles (images) as guidance, which limits their application. Although CLIP-NeRF~\cite{wang2022clip} allows text-guided appearance editing, it only achieves color transfer based on text and cannot transfer stylized details to 3D scenes.  Furthermore, it is a key issue to generate similar results (color tone, details, strokes, etc.) guided by multiple modalities of a style.

In this work, we aim to address the above three problems. We study using multimodal styles as the guidance of 3D style transfer and improving the high-quality rendering by tackling the problems of inconsistent multi-view supervision, style aliasing in multi-style learning, and inconsistency between multimodal guidance.
We propose a new framework of Multimodal-guided 3D Multi-style transfer of NeRF, termed MM-NeRF. In MM-NeRF, to improve the quality of the multi-style transfer, we propose: 1) a novel Multi-view Style Consistent Loss (MSCL) to alleviate the problem of inconstant multi-view supervision, 2) a novel Multi-head Learning Scheme (MLS) to overcome style aliasing caused by the gap among multiple styles.
MSCL specifies one view as the reference view and requires other views to have the same details, further ensuring the consistency of multi-view details in the supervision.
MLS includes a shared backbone for styles to learn common features and independent prediction heads to focus on learning the tones and strokes of each style, thereby reducing style aliasing caused by significant gaps among styles.
For stylization with multimodal guidance, MM-NeRF encodes multimodal styles into a unified feature space and extracts features to guide the stylization.
To keep the style consistent with multimodal guidance, we propose a cross-modal feature correction module to predict feature correction vectors and reduce the distance of cross-modal styles in the feature space.

To the best of our knowledge, MM-NeRF is the first framework that achieves the multimodal-guided 3D multi-style transfer of NeRF. Extensive experiments on several real-world datasets show that MM-NeRF outperforms prior methods and exhibits advantages in multi-view consistency and style consistency of multimodal guidance. Our contributions can be summarized as follows:
\begin{itemize}
    \item 
    We propose MM-NeRF, the first unified framework of multimodal-guided 3D multi-style transfer of NeRF, which achieves high-quality 3D multi-style stylization with multimodal guidance.
    
    \item We analyze problems of lacking details and limited matching with the reference styles in existing 3D stylization methods and reveal that it is due to the inconsistency of multi-view supervision and the style aliasing in multi-style learning. We propose a novel Multi-view Style Consistent Loss (MSCL) and a novel Multi-head Learning Scheme (MLS) to address these problems, which leads MM-NeRF to achieve high-quality stylization.
    
    \item We propose a novel cross-modal feature correction module, which can reduce the distance of cross-modal style in feature space and relieve the problem of inconsistency between multimodal guidance.

    \item Based on MLS, we propose a new incremental learning mechanism, which generalizes the MM-NeRF to any new style with small training costs (a few minutes).    
\end{itemize}

\section{Related Work}


\subsection{Neural Radiance Field}

Given multi-view images of a scene, NeRF \cite{mildenhall2021nerf} and its variants \cite{zhang2020nerf, park2021nerfies, barron2021mip, chen2022sem2nerf, guo2022nerfren, muller2022instant, yang2022recursive, deng2022fov, song2023nerfplayer} can render high quality novel views.
Due to the ability to learn high-resolution complex scenes, NeRF receives widespread attention and has been extended to dynamic scenes \cite{gafni2021dynamic, weng2022humannerf, zhang2022controllable}, large-scale scenes \cite{martin2021nerf, zhang2022nerfusion}, relighting \cite{zhang2021nerfactor, srinivasan2021nerv, rudnev2022nerf}, editing \cite{zhong2023vq}, 
surface reconstruction \cite{wang2021neus},
generation tasks \cite{niemeyer2021giraffe, wang2022clip, chan2022efficient, bergman2022generative}, etc.
In this paper, we focus on 3D style transfer, and NeRF is employed to learn implicit representations of 3D scenes.

\subsection{Neural Style Transfer}

\subsubsection{2D Style Transfer}
Neural image style transfer aims to transfer the style of a reference image to other images using neural networks \cite{gatys2016image}.
For fast image style transfer, methods based on feed-forward networks formulate style transfer as an optimization problem and have made significant progress \cite{johnson2016perceptual, huang2017arbitrary, li2019learning, deng2020arbitrary, svoboda2020two, an2021artflow}.
In recent years, neural style transfer has gradually expanded from images to other modalities \cite{huang2017real, ruder2018artistic, wang2020consistent, chen2020optical, deng2021arbitrary}.

\subsubsection{3D Style Transfer}
Given multi-view images of a scene and a reference style, 3D style transfer aims to synthetic stylized views of this scene with the specified style.
Early 3D style transfer methods are mostly based on point clouds \cite{cao2020psnet, huang2021learning, mu20223d}.
The resolution of 3D scenes limits the stylization quality of these methods.
Recently, NeRF-based 3D style transfer develops rapidly \cite{chen2022upst, zhang2022arf, nguyen2022snerf, chiang2022hyper, huang2022stylizednerf, liu2023stylerf}.
\cite{zhang2022arf} and \cite{nguyen2022snerf} propose high-quality single-style transfer models, but time-consuming optimization is required for each style.
\cite{chiang2022hyper} is the first to formulate the multi-style transfer of NeRF as a parameters prediction problem, which predicts parameters of stylized NeRF with hypernetworks \cite{ha2016hypernetworks}, but their method produces blurry results.
\cite{liu2023stylerf} achieves zero-shot 3D style transfer of arbitrary style by fusing features, but the stylization is limited match with reference styles.
Table \ref{Tb:comparison_3D_style_transfer} compares the applicability of MM-NeRF and other 3D stylization methods.
MM-NeRF achieves high-quality multimodal stylization guided by pre-defined styles and can be generalized to any new style by incremental learning with high-quality stylization and small training costs (a few minutes).

\begin{table*}[t]
    \centering
    \caption{
    Comparison of applicability of 3D style transfer methods. \ding{51} and \ding{55} indicate that the method supports and does not support the specific type of style transfer, respectively. MM-NeRF supports multimodal-guided multi-style transfer and can be generalized to new styles by incremental learning with small training costs (a few minutes).
    }
    \renewcommand\tabcolsep{9pt}
    \begin{tabular*}{\hsize}{@{}@{\extracolsep{\fill}}c c c c c c@{}}
    \hline

    \hline
    \multicolumn{1}{c}{Method} & \multicolumn{1}{c}{Single-Style} & \multicolumn{1}{c}{Multi-Style} & \multicolumn{1}{c}{New Style} & \multicolumn{1}{c}{Guided with Image} & \multicolumn{1}{c}{Guided with Text}\\
    \hline
    ARF \cite{zhang2022arf} & \ding{51} & \ding{55} & \ding{55} & \ding{51} & \ding{55}\\
    SNeRF \cite{nguyen2022snerf} & \ding{51} & \ding{55} & \ding{55} & \ding{51} & \ding{55}\\
    StylizedNeRF \cite{huang2022stylizednerf} & \ding{51} & \ding{51} & \ding{55} & \ding{51} & \ding{55}\\
    HyperNeRF \cite{chiang2022hyper} & \ding{51} & \ding{51} & \ding{51} & \ding{51} & \ding{55}\\
    StyleRF \cite{liu2023stylerf} & \ding{51} & \ding{51} & \ding{51} & \ding{51} & \ding{55}\\
    CLIP-NeRF \cite{wang2022clip} & \ding{51} & \ding{55} & \ding{55} & \ding{51} & \ding{51}\\
    \hline
    MM-NeRF (Ours) & \ding{51} & \ding{51} & \ding{51} & \ding{51} & \ding{51}\\
    
    \hline
    \end{tabular*}
    \label{Tb:comparison_3D_style_transfer}
\end{table*}

\subsection{Multimodal Learning} 

Multimodal representation learning \cite{pan2016jointly, alayrac2020self, ging2020coot, wang2022clip} expands the application scope of generative models \cite{li2018video, rombach2022high, saharia2022photorealistic, chen2017deep, hao2018cmcgan, zhou2018visual}.
By fusing complementary information from multimodal, the performance of related tasks also can be significantly improved.
\cite{li2018video} generates videos with text descriptions.
\cite{rombach2022high} and \cite{saharia2022photorealistic} achieve high-quality image generation based on the Diffusion Model with text.
In the field of 3D generation, current multimodal works mainly focus on text-guided 3D object generation and scene editing \cite{wang2022clip, wei2022hairclip, wang2022nerf, li20223ddesigner, poole2022dreamfusion}. 3D style transfer with multimodal guidance has not been studied yet.
In this work, we propose the first unified framework to achieve multimodal-guided 3D style transfer and tackle the problem of inconsistency between multimodal guidance.

\section{Preliminaries}\label{sec:preliminaries}

NeRF \cite{mildenhall2021nerf} is proposed to model a scene as a combination of opacity and radiation fields.
Given 3D position $\bm{x}\in\mathbb{R}^3$ and viewing direction $\bm{d}\in\mathbb{R}^2$, NeRF predicts opacity $\sigma \in \mathbb{R}^{+}$ and view-dependent radiance color $\bm{c} \in \mathbb{R}^3$.
Ray marching is used to visualize NeRF, and the color of a pixel $C(\bm{r})$ is determined by the integral along the ray $\bm{r}$:
\begin{equation}
\small
 C(\bm{r}) = \int_{t=0}^\infty\tau(t) \sigma(t)\bm{c}(t)dt, \ \tau(t)=e^{-\int_{z=0}^t\sigma(z)dz}
 \label{eq:RayMarching}
\end{equation}
where $t$ is the sampled depth.
NeRF is optimized by minimizing losses between predicted colors and ground truth.
After training, given a new viewpoint, NeRF synthesizes novel views of the same scene.

\begin{figure*}[!ht]
    \centering
    \includegraphics[width=0.99\textwidth]{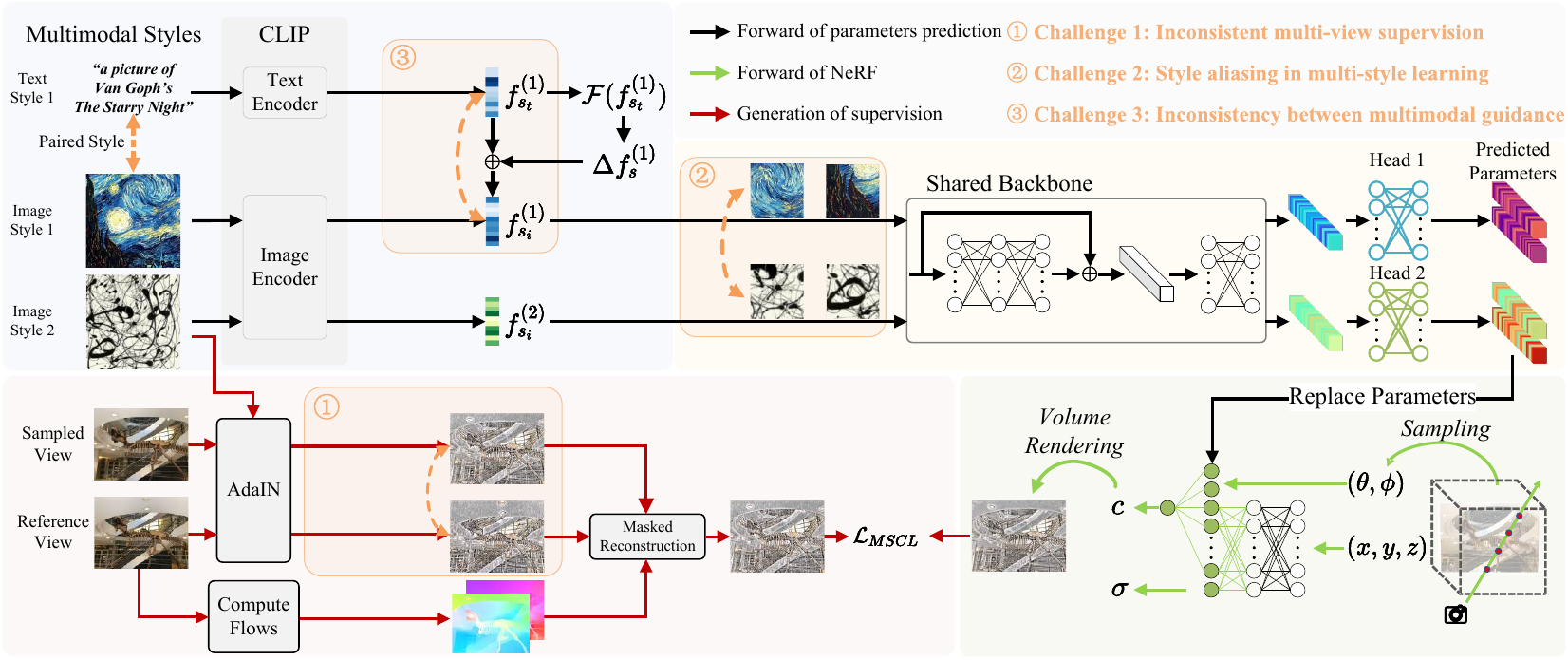}
    \caption{
    The framework of MM-NeRF.
    To address the three challenges, we propose a Multi-view Style Consistent Loss $\mathcal{L}_{MSCL}$, a Multi-head Learning Scheme (MLS), and a novel cross-modal feature correction module $\mathcal{F}(.)$, respectively.
    Multimodal styles are first encoded into CLIP \cite{radford2021learning} space to generate style features, and $\mathcal{F}(.)$ predicts a correction vector $\Delta f_s^{(1)}$ of the feature of text style $f_{s_t}^{(1)}$ to ensure style consistency of multimodal guidance.
    MLS uses a shared backbone to learn common features and predicts the parameters according to the style feature by an independent prediction head. 
    By replacing the color head and the opacity head of the NeRF with the predicted parameters, MM-NeRF renders high-quality stylized images by ray marching.
    $\oplus$ and AdaIN \cite{huang2017arbitrary} are sum and 2D stylization method, respectively.
    }
    \label{fig:main}
\end{figure*}

\section{Method}

\subsection{Problem Formulation}\label{sec:formulation}

\subsubsection{Multimodal-Guided 3D Multi-Style Transfer}
The goal of multimodal-guided 3D multi-style transfer is to synthesize stylized views of scenes with multimodal styles.
In subsequent discussions, we focus on the stylization of NeRF, and this task can be described as follows:
\begin{equation}
\small
\begin{split}
    C(\bm{r}; s^{(i)}) = \int_{t=0}^\infty \tau(t;s^{(i)})\sigma_s(t; s^{(i)})\bm{c}_s(t; s^{(i)})dt, \\ \tau(t;s^{(i)})=e^{-\int_{z=0}^t\sigma_s(z; s^{(i)})dz},s^{(i)} \in S,
    \label{eq:DefinitionOfMM}
\end{split}
\end{equation}
where $S$ is the set of styles, $s^{(i)}$ is the $i$-th style in $S = \{s^{(i)}\}_{i=1}^{N_s}$, $N_s$ is the number of styles in $S$, $\sigma_s$ is the stylized opacity field and $\bm{c}_s$ is the stylized radiation field.
The main task of the multimodal-guided 3D multi-style transfer of NeRF is to design and optimize $\sigma_s$ and $\bm{c}_s$.
The optimization of $\sigma_s$ remains a topic of ongoing debate. Some studies~\cite{huang2022stylizednerf} focused solely on optimizing $\bm{c}_s$, but subsequent researches~\cite{zhang2022arf, nguyen2022snerf} have indicated that simultaneous optimization of $\sigma_s$ and $\bm{c}_s$ yields superior stylization results. In this work, we adopt the method that simultaneously optimizes $\sigma_s$ and $\bm{c}_s$.

\subsubsection{The Challenge of Inconsistent Multi-View Supervision}

Existing 3D style transfer methods are mainly optimized with single-view image-based style loss functions.
In particular, stylized images and features of views, which are used as training data, are generated by 2D style transfer models and 2D encoder models, respectively.
However, these 2D methods do not consider consistency across multiple views and cannot represent spatial information of 3D scenes, which leads to inconsistent stylized multi-view supervision.
As shown in Figure \ref{fig:motivation}, stylized details of the same object are view-dependent and inconsistent across multiple views. NeRF tends to smooth details during stylization training to reduce the total loss of all training views. In this work, we propose a Multi-view Style Consistent Loss (MSCL, Section \ref{sec:MSCL}), which specifies one view as a reference and warps its details to other views to enforce consistency across multi-view supervision.

\subsubsection{The Challenge of Style Aliasing in Multi-Style Learning} Multi-style transfer requires models to generate high-quality style representations for each style. However, as shown in the figure \ref{fig:motivation}, each style has unique tones, strokes, and details, which makes it difficult to represent each style in multi-style learning independently and further leads to style aliasing and limited matching with reference styles (Figure \ref{fig:first_fig} (a)).
In this work, we propose a Multi-head Learning Scheme (MLS, Section \ref{sec:MLS}), which uses a shared backbone to learn common features of styles and declares independent prediction heads to focus on learning each style, thereby reducing style aliasing.

\subsubsection{The Challenge of Inconsistency Between Multimodal Guidance}
Style transfer with multiple modalities of the same style should be able to achieve similar stylized results.
Owing to the inherent ambiguity of textual information, the alignment between textual content and accompanying images lacks a clear and unique correspondence, thereby resulting in inconsistencies within multimodal styles. A more detailed analysis is in Section \ref{sec:MST}. In this work, we propose a novel cross-modal feature correction module to reduce the distance of cross-modal style in feature space and relieve this problem.

\subsection{MM-NeRF}\label{sec:Framework}

The framework of MM-NeRF is illustrated in Figure \ref{fig:main}.
Each style $s^{(i)} \in S$ is encoded into a unified feature space and obtains style feature $f_s^{(i)}$. 
The features of a text style and an image style are denoted as $f_{s_t}^{(i)}$ and $f_{s_i}^{(i)}$, respectively. The Multi-head Learning Scheme predicts the parameters of the stylized color head and the opacity head of NeRF according to style features. After replacing the corresponding parameters of the NeRF with predicted parameters, stylized novel views can be synthesized by Equation (\ref{eq:DefinitionOfMM}).

\subsubsection{Multi-View Style Consistent Loss}\label{sec:MSCL}

Multi-view Style Consistent Loss (MSCL) specifies one view as the reference view and requires others to have the same details. In particular, each stylized training image is computed as follows:
\begin{equation}
    \small
    \label{eq:MSCL}
    I^{\prime(j)} = (1 - M^{(j,ref)}) I^{(j)} + M^{(j,ref)} W^{(j,ref)}(I^{(ref)}),
\end{equation}
where $I^{(j)}$ and $I^{(ref)}$ are the $j$-th stylized image and the reference stylized image with pre-trained AdaIN~\cite{huang2017arbitrary}, respectively, $I^{\prime(j)}$ is the reconstructed stylized image, $W^{(j, ref)}(.)$ is the warping function that warps the reference view to the $j$-th view, and $M^{(j, ref)}$ is the mask.
MSCL warps stylized details of the reference view to other views to ensure that details in the $j$-th view are the same as the reference view. $M^{(j, ref)}$ and $W^{(j, ref)}(.)$ are based on optical flows, which are calculated by RAFT \cite{teed2020raft} with original photo-realistic multi-view images.
Please refer to the appendix for a detailed implementation of generating masks.

In each training step, a mini-batch of rays $\{\bm{r}^{(z)}\}_{z=1}^{M}$ is sampled to compute MSCL, where $M$ is the batch size. 
MSCL is calculated as follows:
\begin{equation}
\small
\label{eq:StylizationTrainingLoss}
\mathcal{L}_{MSCL} = \frac{1}{M}\sum_{z=1}^{M}\Vert \hat{C}(\bm{r}^{(z)}; s^{(i)}) - I^{\prime(j,i)}(\bm{r}^{(z)}) \Vert^2_2,
\end{equation}
where $s^{(i)}$ is the $i$-th style, $\hat{C}(\bm{r}^{(z)}; s^{(i)})$ is the predicted color of ray $\bm{r}^{(z)}$ and $I^{\prime(j,i)}$ is the reconstructed stylized the $j$-th view with the $i$-th style.

\subsubsection{Multi-Head Learning Scheme}\label{sec:MLS}

As shown in Figure \ref{fig:first_fig} (a), the vanilla parameter prediction method results in blurry details and limited matching style with the reference styles.
We propose the Multi-head Learning Scheme (MLS) to address the challenge of style aliasing in multi-style learning. 
MLS consists of a shared backbone and independent prediction heads: 
\begin{equation}
    \small
    \label{eq:MLS}
    \bm{p}^{(i)} = H^{(i)}(B(f_s^{(i)})),
\end{equation}
where $\bm{p}^{(i)}$ the predicted parameters guided by the $i$-th style, $H^{(i)}(.)$ is the prediction head for the $i$-th style, and $B(.)$ is the shared backbone. $B(.)$ can learn common features of styles and $H^{(i)}(.)$ can focus on the learning of the $i$-th style. Compared to the vanilla method, MLS dynamically increases parameters to learn each style, thereby better-addressing style aliasing caused by multiple styles.
To reduce the number of parameters to be predicted, MM-NeRF employs MLS to predict the parameters of two prediction heads of the NeRF rather than all parameters.

\begin{figure}[t]
    \centering
    \includegraphics[width=0.99\linewidth]{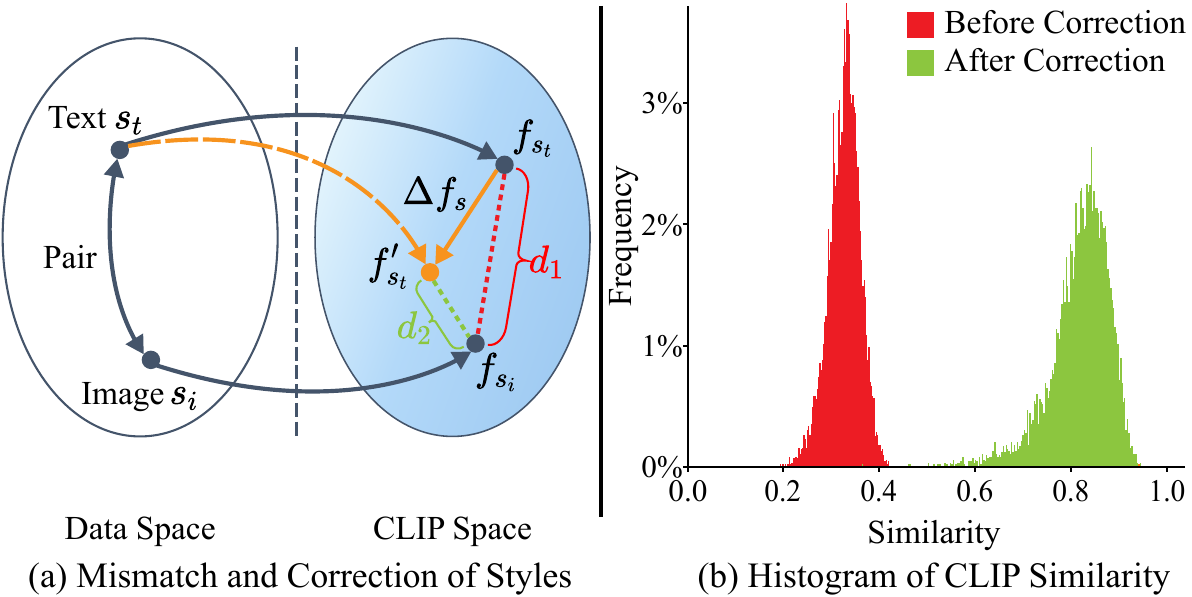}
    \caption{
    (a) Illustration of inconsistency between multimodal guidance. A pair of text $s_t$ and an image $s_i$ are mapped to $f_{s_t}$ and $f_{s_i}$, resulting in mismatched stylization. MM-NeRF addresses this problem by introducing a cross-modal feature correction module to predict the feature correction vector $\Delta f_s$ and corrects the feature of the text from $f_{s_t}$ to $f_{s_t}^\prime$. The distance between features of the text-image pair is reduced from $d_1$ to $d_2$. (b) Histogram of CLIP similarity of text-image pairs before and after correction.
    }
    \label{fig:multimodal_consistency}
\end{figure}

\subsubsection{Cross-Modal Feature Correction Module}\label{sec:MST}

One solution to adopt multimodal guidance is to introduce semantic supervision, such as CLIP-NeRF \cite{wang2022clip}, achieving conditional generation guided by text with CLIP loss. However, semantic losses make it challenging to represent style strokes. In addition, due to the high demand for the memory of NeRF, only patch-based training is usually possible, leading to low-quality stylization. 
Therefore, for each text style, we generate an image with Stable Diffusion \cite{rombach2022high} and construct a text-image pair, where the text is used to guide style transfer, and the image is used to generate supervision.

The styles of two modalities in any image-text pair should be encoded to the same style features to ensure style consistency of multimodal guidance.
However, the text and the image are mapped to different style features (Figure \ref{fig:multimodal_consistency} (a)). The CLIP similarity of text-image pairs is also extremely low (Figure \ref{fig:multimodal_consistency} (b)).
There are two reasons for this situation: 1) the ambiguity of the text and 2) the inherent bias of the text-to-image model. The ambiguity of the text results in non-unique mapping relationships between image and text. The inherent bias of the text-to-image model leads to differences between the generated image and the ground truth (saturation, brightness, etc.), even if there is a unique corresponding image-text pair.
MM-NeRF addresses this problem with a novel cross-modal feature correction module $\mathcal{F}(.)$ to predict the feature correction vector $\Delta f_s$. The style feature of a style $s$ is as follows:
\begin{equation}
\small
    f_s = \left\{\begin{array}{ll}
         E_i(s), \ &if \ s \ is \ an \ image,  \\
         E_t(s) + \mathcal{F}(s), \ &else,
    \end{array}\right.
    \label{eq:MST}
\end{equation}
where $E_i(.)$ and $E_t(.)$ are the image encoder and the text encoder of CLIP \cite{radford2021learning}, respectively.
As illustrated in Figure \ref{fig:multimodal_consistency}, the cross-modal feature correction module can significantly improve the consistency of multimodal style.

\subsubsection{Learning}\label{sec:Learning}

The stylization training is shown as Algorithm \ref{alg:Training}.
We first pre-train basic NeRF with multi-view images and fix it in the stylization training.
The cross-modal feature correction module is pre-trained with CLIP similarity loss and mean-squared loss on text-image pairs generated by Stable Diffusion \cite{rombach2022high}. See the Appendix for additional pre-train details.
Multi-view images are denoted as $\{I_c^{(j)}\}_{j=1}^{N_c}$, where ${I_c^{(j)}}$ is the $j$-th image and $N_c$ is the number of training images of this scene.
At each iteration of stylization training, we randomly sample an image $I_c$ from $\{I_c^{(j)}\}_{j=1}^{N_c}$ and a style $s^{(i)}$ from the set of styles $S$ to compute the reconstructed stylized image $I^{\prime(j, i)}$.
Parameters of the two prediction heads of the NeRF are predicted as described in Section \ref{sec:MLS}, and NeRF predicts the stylized color of pixels by ray marching.
MLS is optimized to minimize MSCL.

\begin{algorithm}[tb]
\caption{Stylization Training of MM-NeRF}
\label{alg:Training}
\textsc{Input}: Style set $S$, multi-view images of a real-world scene $\{I_c^{(i)}\}_{i=1}^{N_c}$, NeRF pre-trained on $\{I_c^{(i)}\}_{i=1}^{N_c}$, pre-trained cross-modal feature correction module, the Multi-head Learning Scheme (MLS). \\
\textsc{Parameter}: Iteration steps of stylization training $T_s$, batch size of each step $M$. \\
\textsc{Stylization Training}
\begin{algorithmic}[1]
\FOR{iteration $t = 1, \cdots, T_s$}
\STATE Randomly sample an image $I_c^{(j)}$ from $\{I_c^{(j)}\}_{j=1}^{N_c}$ and a style $s^{(i)}$ from $S$.
\STATE Extract the feature vector $f_s^{(i)}$ according to Equation \ref{eq:MST}.
\STATE Predict parameters $\bm{p}^{(i)}$ according to Equation \ref{eq:MLS}.
\STATE Compute reconstructed stylized image $I^{(j,i)}$ according to Equation \ref{eq:MSCL}.
\STATE Randomly sample a mini-batch $\{\bm{r}^{(z)}\}_{z=1}^M$ of rays.
\STATE Predict colors $\hat{C}(\bm{r}^{(z)}; s^{(i)})$ according to Equation \ref{eq:DefinitionOfMM}.
\STATE Optimize MLS to minimize MSCL (Equation \ref{eq:StylizationTrainingLoss}).
\STATE Update the parameters of MLS.
\ENDFOR
\end{algorithmic}
\end{algorithm}

\subsection{Incremental Learning Scheme of New Styles}\label{sec:IncrementalLearning}

\begin{figure*}[t]
    \centering
    \includegraphics[width=0.99\textwidth]
    {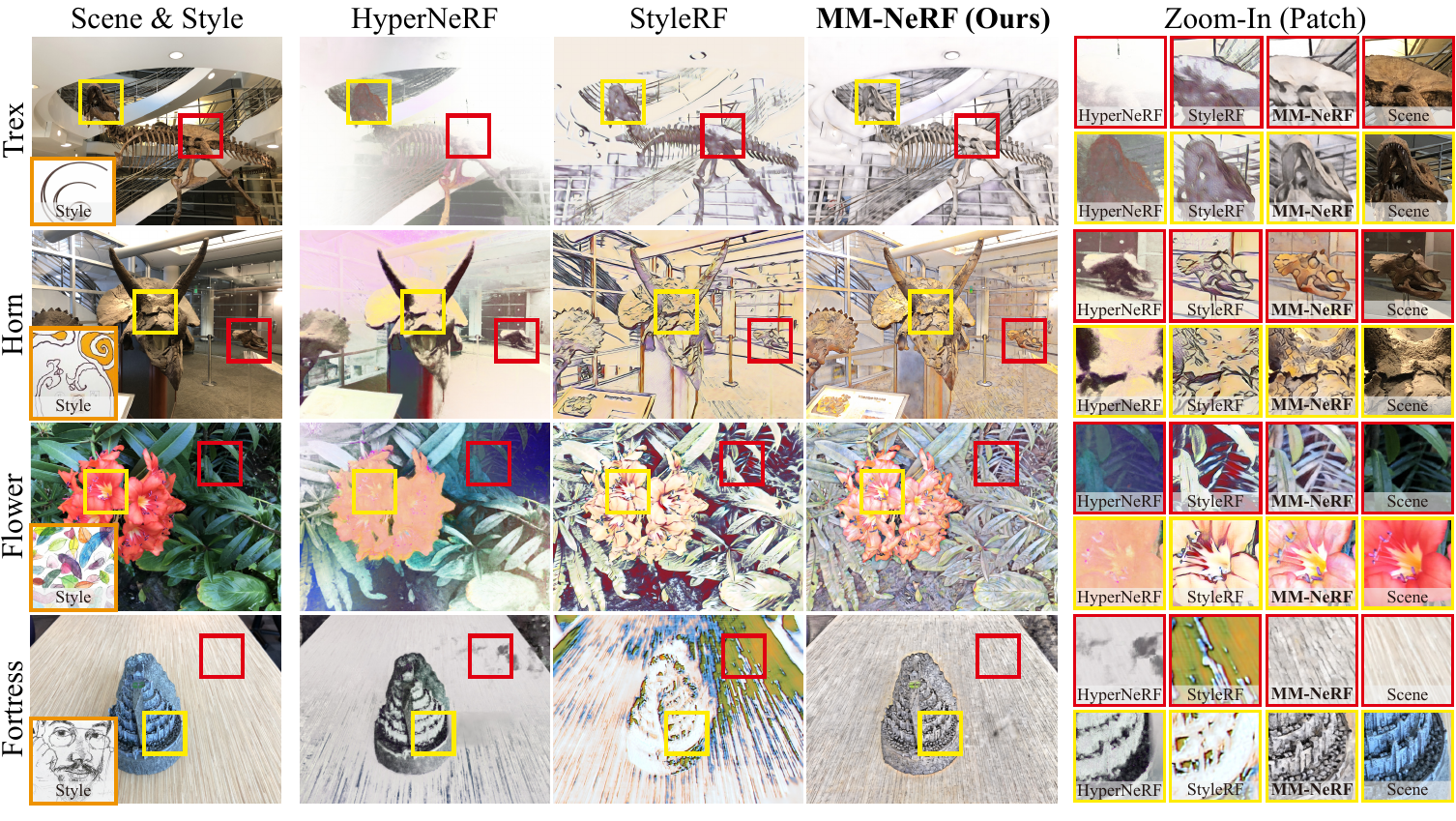}
    \caption{
    Comparison of image-guided 3D style transfer with HyperNeRF \cite{chiang2022hyper} and StyleRF \cite{liu2023stylerf} on forward-facing scenes \cite{mildenhall2021nerf}.
    MM-NeRF generates high-quality stylized images with texture details.
    }
    
    \label{fig:main_comparison}
\end{figure*}

\begin{figure*}[t]
    \centering
    \includegraphics[width=0.99\textwidth]{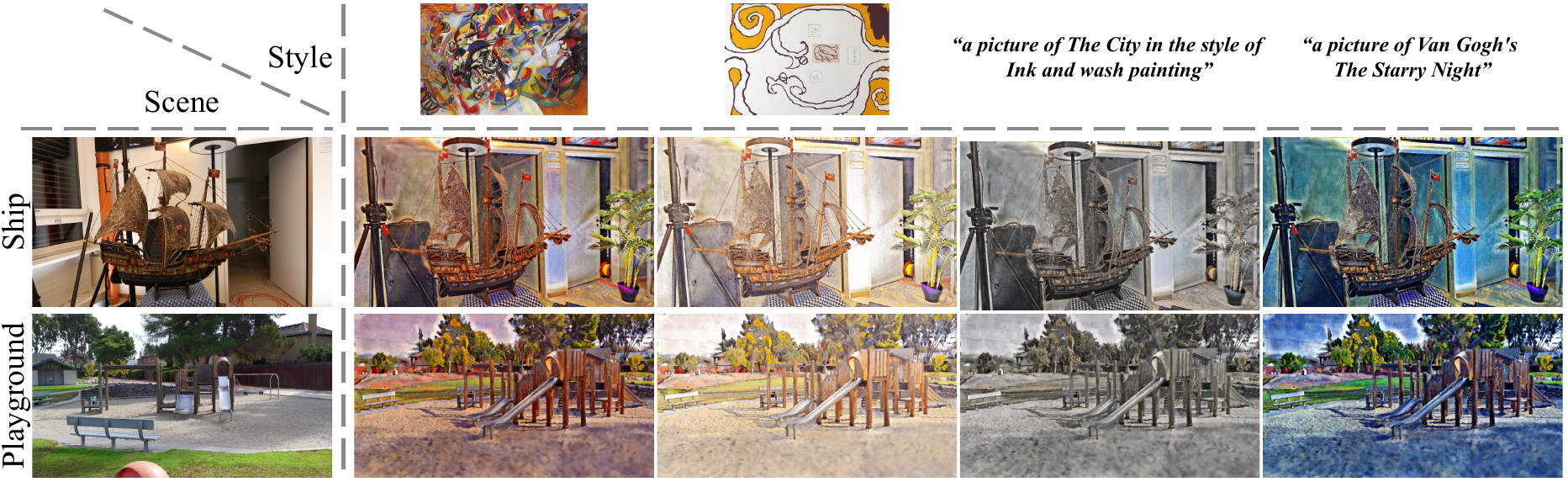}
    \caption{Multimodal 3D style transfer. MM-NeRF achieves multimodal-guided 3D style transfer on ship \cite{yucer2016efficient} and playground \cite{knapitsch2017tanks}.
    }
    
    \label{fig:mutlimodal-guided}
\end{figure*}

\subsubsection{Similarity Measurement}
For a new style $s^{(new)}$, we first measure similarity between $s^{(new)}$ and other styles in the style set $S$ on feature space by cosine distance:
\begin{equation}
    \small
    \bm{f}_{s^{(sim)}} = \min_{\bm{f}_{s^{(i)}}} (1- \frac{\bm{f}_{s^{(i)}}^T \cdot \bm{f}_{s^{(new)}}}{\Vert \bm{f}_{s^{(i)}} \Vert_2 \cdot \Vert \bm{f}_{s^{(new)}} \Vert_2}), s^{(i)} \in S,
    \label{eq:FeatureSim}
\end{equation}
where $\bm{f}_{s^{(new)}}$ and $\bm{f}_{s^{(i)}}$ are the feature vectors of $s^{(new)}$ and $s^{(i)}$, respectively. $s^{(sim)}$ is the most similar style to the new style in the style set, and $\bm{f}_{s^{(sim)}}$ is the feature vector of $s^{(sim)}$.
If the cosine distance between $s^{(sim)}$ and $s^{(new)}$ is smaller than the threshold, it signifies that the style is already included in the style set, prompting the invocation of the prediction head of $s^{(sim)}$ for the stylization.
Otherwise, incremental learning for style ${s^{(new)}}$ will be performed.

\subsubsection{Incremental Learning}
A new prediction head is initialized with the prediction head of $s^{(sim)}$ for a new style.
The incremental learning of new styles is similar to the stylization training described in Section \ref{sec:Learning}.
The only difference is that the shared backbone of MLS is fixed during incremental learning.
In particular, as only new prediction heads need to be optimized, incremental learning of new styles can be finished in several minutes.

\section{Experiments}

\subsection{Experiment Settings}

\subsubsection{Datasets}
We conduct extensive experiments on several real-world 3D scene datasets, including forward-facing scenes \cite{mildenhall2021nerf}, Tanks and Temples \cite{knapitsch2017tanks}, and 360° scenes \cite{yucer2016efficient}.
Images from WikiArt \cite{painter-by-numbers} and self-written texts are used to guide style transfer.

\subsubsection{Metrics}
3D style transfer models are evaluated on multi-view consistency and stylized quality.
In our experiments, we follow the usual practice to evaluate multi-view consistency by Temporal Warping Error (TWE) and warped Learned Perceptual Image Patch Similarity (LPIPS) \cite{huang2021learning, chiang2022hyper, nguyen2022snerf}.
In particular, we report the average results of each dataset guided by five styles.
Stylization quality is mainly evaluated by a user study.

\subsubsection{Baselines}
As MM-NeRF can be generated to new styles, we compare MM-NeRF to State-Of-The-Art (SOTA) NeRF-based arbitrary style transfer methods, including StyleRF \cite{liu2023stylerf} and HyperNeRF \cite{chiang2022hyper}.
Furthermore, the methodology of generating novel views with NeRF \cite{mildenhall2021nerf} and subsequently employing 2D style transfer methods~\cite{huang2017arbitrary, deng2022stytr2, deng2021arbitrary} for stylization serves as the baseline in our experiments.
Specifically, we employ AdaIN~\cite{huang2017arbitrary}, StylTR$^2$~\cite{deng2022stytr2} and MCCNet~\cite{deng2021arbitrary} to implement baselines. AdaIN~\cite{huang2017arbitrary} and StyTR$^2$~\cite{deng2022stytr2} are SOTA image style transfer methods.
MCCNet~\cite{deng2021arbitrary} is a SOTA video style transfer method.
Since MM-NeRF is the first multimodal-guided 3D style transfer model, we mainly compare MM-NeRF with these methods on image-guided 3D style transfer.

\begin{figure*}[!ht]
    \centering
    \includegraphics[width=0.99\textwidth]{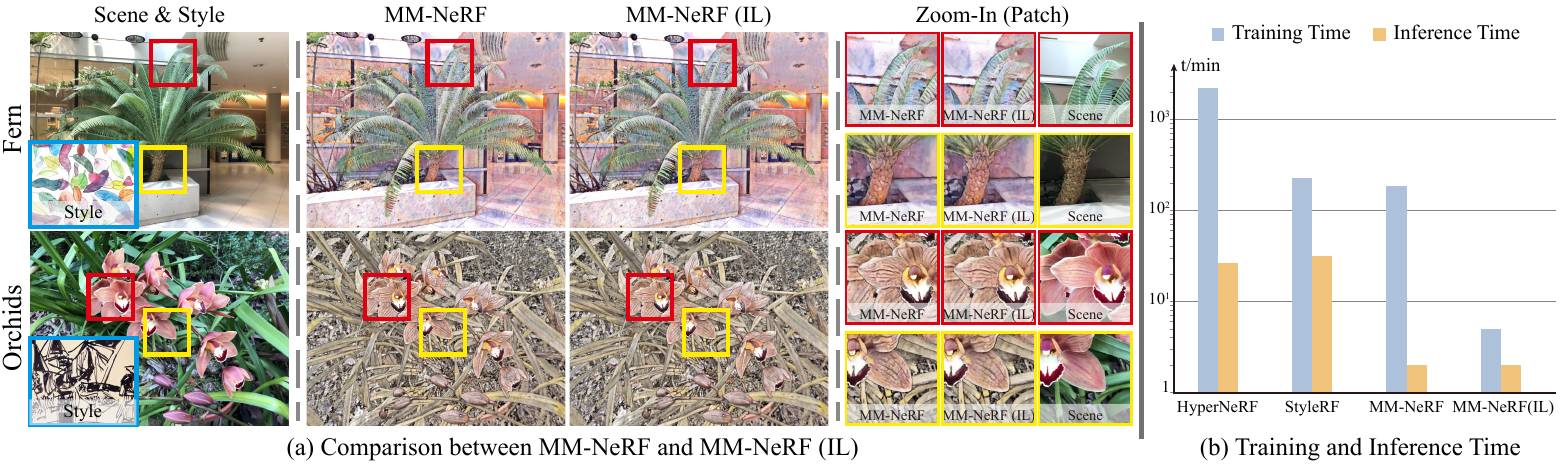}
    \caption{
    Incremental learning of MM-NeRF. (a) Incremental learning on forward-facing scenes \cite{mildenhall2021nerf}. (b) Comparison of stylization training time and inference time. The training time is the average time for each scene, and the inference time is the average time for rendering 120 frames.
    }
    \label{fig:incremental_learning}
\end{figure*} 

\begin{figure}[t]
    \centering
    \includegraphics[width=0.99\linewidth]
    {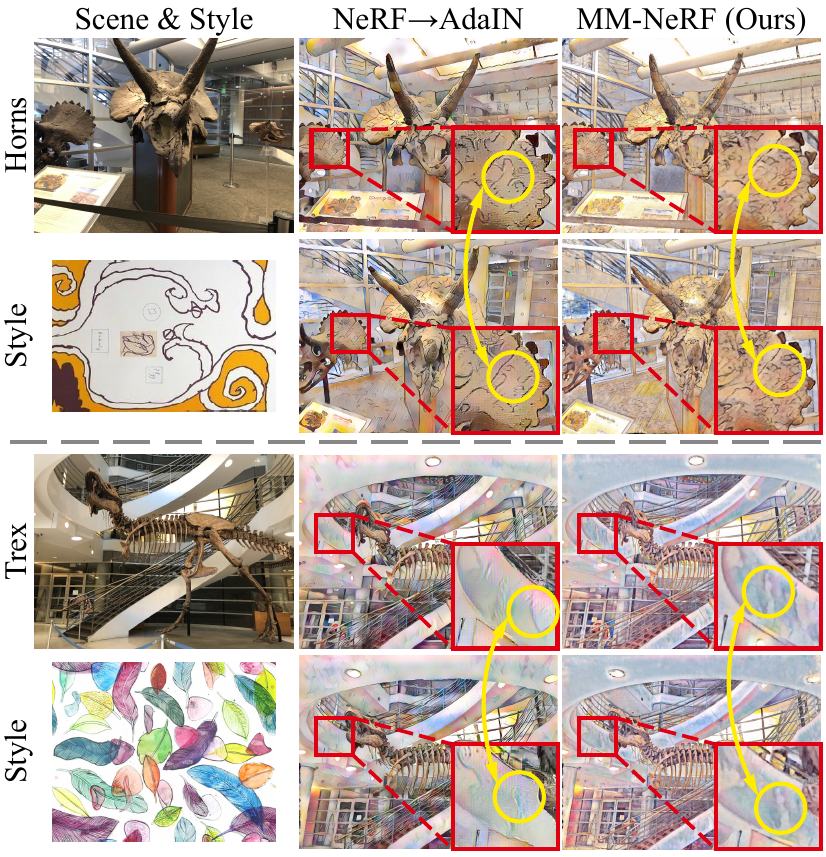}
    \caption{
    MM-NeRF can ensure consistency of multi-view.
    }
    
    \label{fig:multiview_consistency}
\end{figure}

\begin{figure}[t]
    \centering
    \includegraphics[width=0.99\linewidth]
    {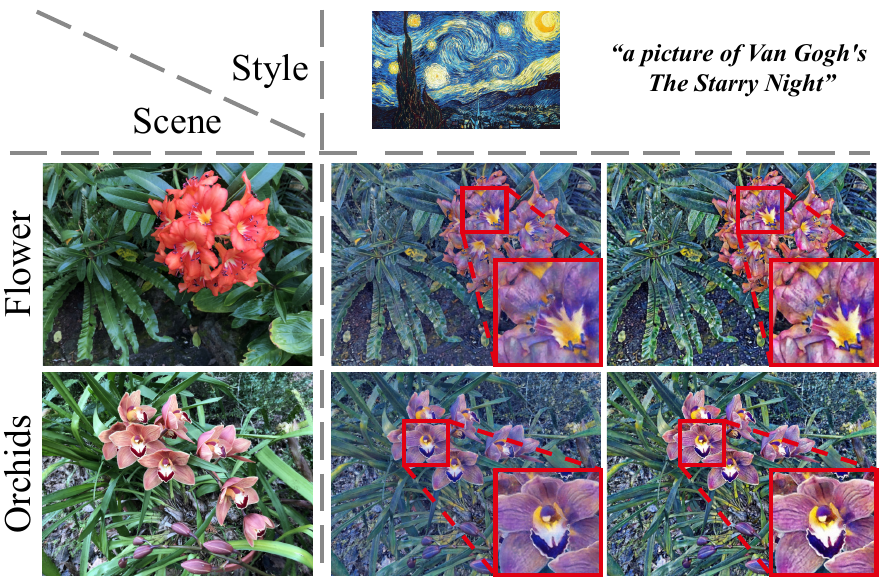}
    \caption{
    MM-NeRF achieves high-quality stylization with multimodal styles.}
    
    \label{fig:multimodal_comparison}
\end{figure}

\subsubsection{Training Configurations}
For each scene, NeRF is pre-trained for 30k iterations, and MM-NeRF is trained for 50k iterations on a single NVIDIA GTX 4090 GPU. We pre-generate all reconstructed stylized images according to Equation \ref{eq:MSCL} for each scene to improve training efficiency.

\begin{table}[t]
    \centering
    \caption{
    User study. MM-NeRF outperforms SOTA methods on multi-view consistency and stylization quality.
    }
    \label{Tb:user_study}
    \renewcommand\tabcolsep{9pt}
    \begin{tabular*}{\hsize}{@{}@{\extracolsep{\fill}}c c c c@{}}
    \hline

    \hline
    \multicolumn{1}{c}{\multirow{2}{*}{Method}} & \multicolumn{1}{c}{\multirow{2}{*}{Consistency($\uparrow$)}} & \multicolumn{2}{c}{Stylization}\\
    \cline{3-4} & & Overall($\uparrow$) & Detail($\uparrow$)\\
    \hline
    StyleRF \cite{liu2023stylerf} & 3.7 & 3.2 & 2.7\\
    HyperNeRF \cite{chiang2022hyper} & 2.8 & 1.9 & 1.9\\
    \hline
    MM-NeRF (Ours) & 4.6 & 4.1 & 4.0\\
    
    \hline
    \end{tabular*}
\end{table}

\subsection{Main Results}\label{sec:main_compare}

\subsubsection{High-Quality 3D Multi-Style Transfer}
In Figure \ref{fig:main_comparison}, we compare MM-NeRF with StyleRF and HyperNeRF on forward-facing scenes \cite{mildenhall2021nerf}.
MM-NeRF generates high-quality stylized novel views with texture details.
By contrast, stylized images with other methods show blurry details and limited matching with the reference styles.
In particular, the results of HyperNeRF exhibit a significant loss of details, making it difficult to discern objects within the scene. Similarly, StyleRF also suffers from a lack of details and tends to produce high-contrast results that may cause discomfort. In comparison, the results generated by MM-NeRF not only adhere to the style defined by the reference images but also exhibit clear details.

To better evaluate MM-NeRF, StyleRF, and HyperNeRF, we conducted a user study. 
As shown in Table \ref{Tb:user_study}, the stylization of MM-NeRF outperforms other methods. The detailed analysis of the user study is presented in Section \ref{Sec:UserStudy}.

\subsubsection{Multimodal-Guided 3D Style Transfer}
Figure \ref{fig:mutlimodal-guided} reports multimodal-guided 3D style transfer of MM-NeRF on Tanks and Temples \cite{knapitsch2017tanks} and 360° scenes \cite{yucer2016efficient}.
Figure \ref{fig:multimodal_comparison} reports multimodal-guided 3D style transfer of MM-NeRF on forward-facing scenes.
MM-NeRF can generate stylized novel views with reasonable color tones and details with the guidance of images and text.

\subsubsection{Multi-View Consistency}
Figure \ref{fig:multiview_consistency} compares the multi-view consistency with the baseline.
MM-NeRF can effectively ensure that the same objects have similar details in multiple views.
Table \ref{Tb:metric_llff} reports qualitative results of multi-view consistency on forward-facing scenes \cite{mildenhall2021nerf}. 
As it is hard for 2D style transfer methods to ensure multi-view consistency, 3D style transfer methods significantly outperform baselines.
Furthermore, MM-NeRF outperforms other SOTA 3D stylization methods.
MM-NeRF outperforms StyleRF by 20.5\% and 3.0\% in TWE and warped LPIPS, respectively.
MM-NeRF outperforms HyperNeRF by 19.5\% and 15.1\% in TWE and warped LPIPS, respectively.
Table \ref{Tb:metric_tnt} reports qualitative results of multi-view consistency on Tanks and Temples \cite{knapitsch2017tanks}.
MM-NeRF outperforms HyperNeRF by 14.0\% and 1.3\% in TWE and warped LPIPS, respectively.
These results demonstrate that MM-NeRF can effectively ensure multi-view consistency.

\begin{table}[t]
    \centering
    \caption{
    Comparison of multi-view consistency with SOTA methods on forward-facing scenes \cite{mildenhall2021nerf} in TWE($\downarrow$) and warped LPIPS($\downarrow$) as \cite{huang2021learning, chiang2022hyper, nguyen2022snerf}.
    }
    \label{Tb:metric_llff}
    \renewcommand\tabcolsep{9pt}
    \begin{tabular*}{\hsize}{@{}@{\extracolsep{\fill}}c c c@{}}
    \hline

    \hline
    \multicolumn{1}{c}{Method} & \multicolumn{1}{c}{TWE$\times 10^{-2}$($\downarrow$)} & \multicolumn{1}{c}{LPIPS$\times 10^{-2}$($\downarrow$)}\\
    \hline
        NeRF $\rightarrow$ AdaIN & 5.35 & 7.78\\
        NeRF $\rightarrow$ StyTR$^2$~\cite{deng2022stytr2} & 4.37 & 8.28\\
        NeRF $\rightarrow$ MCCNet~\cite{deng2021arbitrary} & 2.34 & 7.88\\
    \hline
        StyleRF \cite{liu2023stylerf} & 1.56 & 7.27\\
        HyperNeRF \cite{chiang2022hyper} & 1.54 & 8.30\\
    \hline
    
    MM-NeRF (Ours) & 1.24 & 7.05\\
    
    \hline
    \end{tabular*}
\end{table}

\begin{table}[t]
    \centering
    \caption{
    Comparison of multi-view consistency with SOTA methods on Tanks and Temples \cite{knapitsch2017tanks} in TWE($\downarrow$) and warped LPIPS($\downarrow$) as \cite{huang2021learning, chiang2022hyper, nguyen2022snerf}. (StyleRF cannot adapt to this dataset.)
    }
    \label{Tb:metric_tnt}
    \renewcommand\tabcolsep{9pt}
    \begin{tabular*}{\hsize}{@{}@{\extracolsep{\fill}}c c c@{}}
    \hline

    \hline
    \multicolumn{1}{c}{Method} & \multicolumn{1}{c}{TWE$\times 10^{-2}$($\downarrow$)} & \multicolumn{1}{c}{LPIPS$\times 10^{-2}$($\downarrow$)}\\
    \hline
        NeRF $\rightarrow$ AdaIN & 4.41 & 12.63\\
        NeRF $\rightarrow$ StyTR$^2$~\cite{deng2022stytr2} & 1.88 & 13.33\\
        NeRF $\rightarrow$ MCCNet~\cite{deng2021arbitrary} & 2.54 & 10.63\\
    \hline
        StyleRF \cite{liu2023stylerf} & - & -\\
        HyperNeRF \cite{chiang2022hyper} & 0.57 & 8.97\\
    \hline
    
    MM-NeRF (Ours) & 0.49 & 8.85\\
    
    \hline
    \end{tabular*}
\end{table}

\subsubsection{Incremental Learning of New Styles}
MM-NeRF can generalize to new styles by Incremental Learning (IL).
To evaluate the quality of new styles learned by IL, we remove a style from the style set and train MM-NeRF with the remaining styles. In addition, the removed style is learned by IL.
As shown in Figure \ref{fig:incremental_learning} (a), MM-NeRF can generate similarly stylized novel views with pre-defined and new styles.
Figure \ref{fig:incremental_learning} (b) reports the comparison of efficiency with StyleRF and HyperNeRF. As the pre-training periods of MM-NeRF, HyperNeRF, and StyleRF are different, we only compare the stylization training time and inference time in Figure \ref{fig:incremental_learning} (b).
MM-NeRF can generalize to new styles with small training costs (a few minutes).
The inference efficiency of MM-NeRF is approximately $\times 55$ faster than SOTA methods.

\begin{figure*}[!ht]
    \centering
    \includegraphics[width=0.95\textwidth]
    {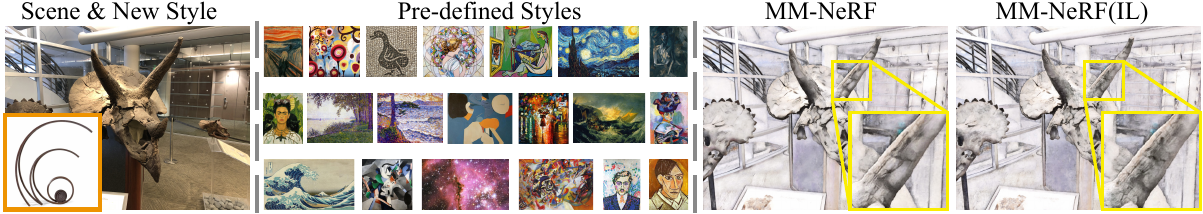}
    \caption{
    Incremental Learning of MM-NeRF for a new style. Although this new style significantly differs from pre-defined styles, MM-NeRF still achieves high-quality stylization.
    }
    
    \label{fig:IL}
\end{figure*}

We design an additional experiment to validate the effectiveness of IL further. In this experiment, MM-NeRF asks to learn a significantly different style by IL after training on pre-defined styles. The results are shown in Figure \ref{fig:IL}.
Compared to the pre-defined learning, the quality of stylized results guided by the new style has only slightly decreased, and MM-NeRF still achieves excellent stylization.

\begin{figure*}[t]
    \centering
    \includegraphics[width=0.99\textwidth]{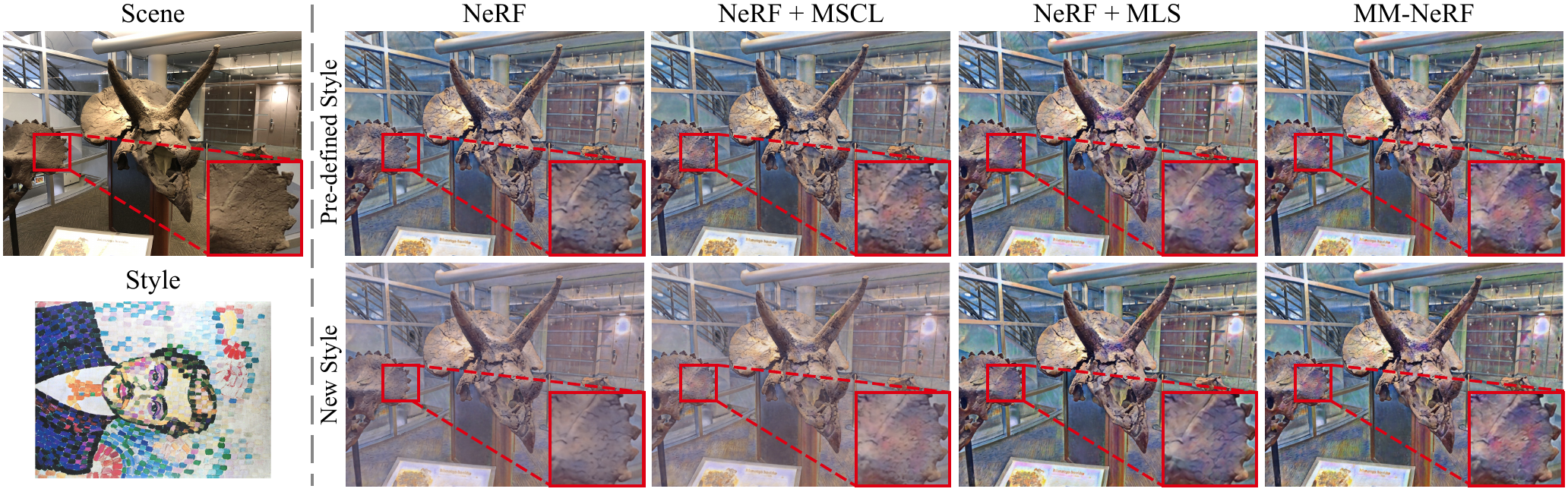}
    \caption{
    Ablation study of MSCL and MLS on forward-facing scenes \cite{mildenhall2021nerf}. MSCL and MLS can improve stylization quality significantly.
    }
    \label{fig:ablation_study}
\end{figure*}

\begin{table}[t]
    \centering
    \caption{
    Ablation study of MSCL and MSL on forward-facing scenes \cite{mildenhall2021nerf} with LPIPS($\downarrow$), PSNR($\uparrow$), and SSIM($\uparrow$). 
    }
    \label{Tb:ablation}
    \renewcommand\tabcolsep{9pt}
    \begin{tabular}{@{}@{\extracolsep{\fill}}c c c c@{}}
    \hline

    \hline
    \multicolumn{1}{c}{Method} & \multicolumn{1}{c}{LPIPS$\times 10^{-2}$($\downarrow$)} & \multicolumn{1}{c}{PSNR($\uparrow$)} & \multicolumn{1}{c}{SSIM$\times 10^{-2}$($\uparrow$)}\\
    \hline

    NeRF & 35.42 & 19.38 & 49.14\\
    \hline
    NeRF + MSCL & 35.59 & 19.60 & 50.53\\
    NeRF + MLS & 29.00 & 19.63 & 51.19\\
    MM-NeRF & 28.33 & 19.92 & 53.83\\
    
    \hline
    \end{tabular}
\end{table}

\subsection{User Study}\label{Sec:UserStudy}

A user study is conducted to compare MM-NeRF with StyleRF and HyperNeRF on multi-view consistency and stylization.
We generate videos and stylized novel views by these stylization methods and ask participants to rate multi-view consistency and stylization on a scale of 1-5 (a higher score represents better multi-view consistency and stylization).
A total of 86 questionnaires were collected, and the results are shown in Table \ref{Tb:user_study}.
MM-NeRF outperforms other methods on all metrics. Specifically, MM-NeRF outperforms StyleRF and HyperNeRF by 24.3\% and 64.3\% on multi-view consistency, respectively.
In addition, MM-NeRF also outperforms SOTA methods on overall stylized quality and details.
Especially in the evaluation of details, MM-NeRF achieves almost twice the score of the SOTA methods.
The user study verifies that MM-NeRF can generate better stylized novel views than StyleRF and HyperNeRF.

\subsection{Ablation Studies}

Ablation studies are conducted on forward-facing scenes.
Specific stylized frame sequences are employed as the ground truth to evaluate the efficiency of the proposed modules.
Specifically, these sequences are generated by AdaIN and corrected via optical flow.

\subsubsection{Multi-View Style Consistent Loss}
As shown in Figure \ref{fig:ablation_study}, unlike Multi-head Learning Scheme (MLS), which enhances stylization quality at a global level, Multi-view Style Consistent Loss (MSCL) primarily enriches the level of detail.
MSCL ensures the consistency of details in multi-view supervised data, thereby mitigating issues of detail blurring.
Table \ref{Tb:ablation} reports the qualitative results of stylized details on forward-facing scenes \cite{mildenhall2021nerf}. NeRF with MSCL outperforms NeRF by 2.8\% in SSIM. In addition, the improvement based on NeRF+MLS is more significant that MM-NeRF outperforms NeRF+MLS by 2.3\%, 1.5\%, and 5.2\% in LPIPS, PSNR, and SSIM, respectively.
Both qualitative and quantitative experiments verified that MSCL effectively enhances the quality of local details in stylized results.

To explore the relationship between the reference view in MSCL and stylization quality, we conduct experiments on \emph{fern} from forward-facing scenes \cite{mildenhall2021nerf}.
Specifically, each view is used as the reference view to implement MSCL and train MM-NeRF independently. After training, we calculate the SSIM for each experiment to evaluate stylization quality.
Since MSCL needs to determine the warp function and masks based on optical flow, we also calculate the ratio of masked pixels in each experiment.

\begin{figure}[t]
    \centering
    \includegraphics[width=0.92\linewidth]
    {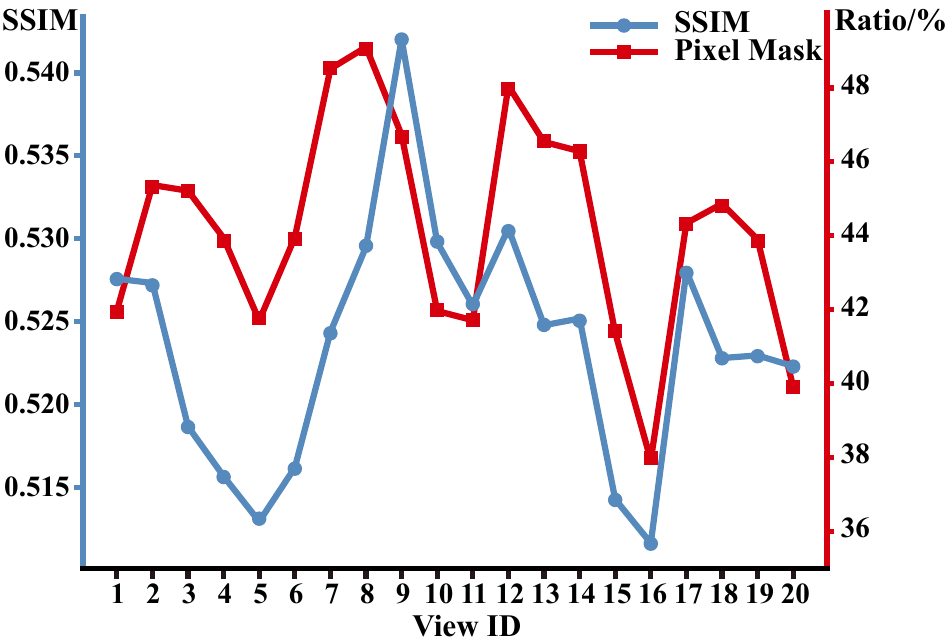}
    \caption{
    The blue line represents the relationship between SSIM and the reference view. The red line represents the relationship between the ratio of masked pixels and the reference view.
    }
    \label{fig:reference_view}
\end{figure}

As shown in Figure \ref{fig:reference_view}, the reference view significantly impacts stylization quality. In addition, there is a significant positive correlation between stylization quality and the ratio of masked pixels.
A larger ratio of masked pixels means the reference view has more pixels that match other views, thereby better ensuring the consistency of multi-view details in supervision and improving stylization quality. Based on the above discussion, we recommend using the view with the maximum ratio of masked pixels as the reference view in MSCL to generate better stylized novel views.

\subsubsection{Multi-Head Learning Scheme}
To verify the effectiveness of MLS, we compare MLS with the vanilla parameter prediction method, in which styles share a prediction head.
As shown in Figure \ref{fig:ablation_study}, MLS generates stylized results with more texture details.
For new styles, MLS can also generate high-quality results.
By contrast, the stylized results by the vanilla parameter prediction method show limited matching with the reference style.
Table \ref{Tb:ablation} reports the results of stylized detail on forward-facing scenes \cite{mildenhall2021nerf}.
NeRF with MLS outperforms NeRF by 18.1\%, 1.3\%, and 4.2\% in LPIPS, PSNR, and SSIM, respectively.
These results demonstrate that MLS can effectively improve the quality of stylization.

\subsubsection{Style Consistency of Multimodal Guidance}
\begin{figure}[t]
    \centering
    \includegraphics[width=0.99\linewidth]{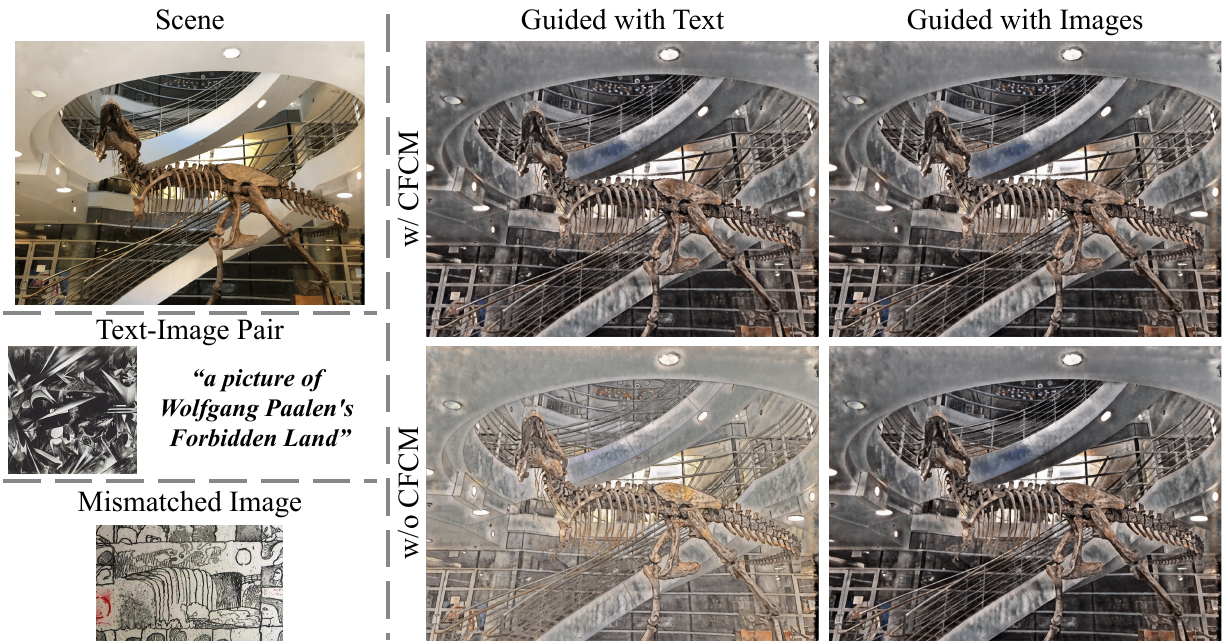}
    \caption{
    Ablation study of Cross-modal Feature Correction Module (CFCM) on forward-facing scenes \cite{mildenhall2021nerf}. MM-NeRF without CFCM mistakenly matches the text to the wrong image.
    }
    \label{fig:semantic_consistency}
\end{figure}

To verify the effectiveness of the cross-modal feature correction module, we conduct a comparative experiment on forward-facing scenes \cite{mildenhall2021nerf}. For a specific text-image pair, we add the image to the pre-defined set of styles and train MM-NeRF with this set. We verify whether the text and the image can be recognized as the same style.
IL is not performed in this experiment, and only similarity measurement is performed.
Figure \ref{fig:semantic_consistency} reports the results that MM-NeRF with the cross-modal feature correction module can effectively recognize the image and the text as the same style, while MM-NeRF without it mistakenly matches the text to a wrong image.

\section{Future Work}

One side effect of MLS is that, as the number of styles increases, more memory is needed to store a large number of prediction heads. Since MLS only predicts parameters and does not participate in rendering new views, these heads do not affect inference efficiency. However, this defect is still a limitation hindering the application of MM-NeRF. We will try to solve this problem in our future work, such as introducing parameter quantification and hierarchical management of all styles according to usage frequency.

In addition, MM-NeRF currently only supports global stylization and is unable to achieve local stylization, which is more helpful for communities, 3D scene visualization, and VR applications. Therefore, we aim to extend MM-NeRF to support local stylization of 3D scenes in our future work.

\section{Conclusion}
In this paper, we propose MM-NeRF, the first unified framework of multimodal-guided 3D multi-style transfer of NeRF.
We reveal three challenges in multimodal-guided 3D multi-style transfer: 1) inconsistent multi-view supervision, 2) style aliasing in multi-style learning, and 3) inconsistency between multimodal guidance. 
To tackle these problems, we propose a novel Multi-view Style Consistent Loss (MSCL) to alleviate the problem of inconstant multi-view supervision, a novel Multi-head Learning Scheme (MLS) to overcome style aliasing, and a novel cross-modal feature correction module to reduce the distance of cross-modal style in feature space to ensure consistency of multimodal guidance.
Furthermore, based on MLS, we propose a new incremental learning mechanism to generalize MM-NeRF to any new styles with small training costs (a few minutes).
Experiments on several real-world datasets demonstrate that MM-NeRF outperforms prior SOTA methods and exhibits advantages in multi-view consistency and style consistency of multimodal guidance.

\section*{Acknowledgments}
We extend our special thanks to the Beijing Engineering Research Center of Industrial Spectrum Imaging and the National Environmental Corrosion Platform of China for their invaluable support of computational resources.

\bibliographystyle{IEEEtran}
\bibliography{ref}


\begin{IEEEbiography}[{\includegraphics[width=1in,height=1.25in,clip,keepaspectratio]{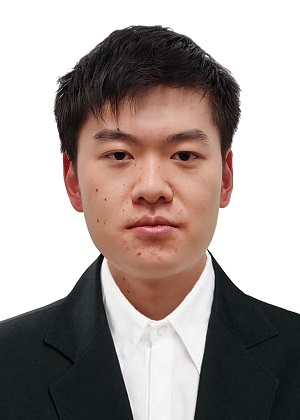}}]{Zijiang Yang}
received the B.S. degree in materials science and engineering from the University of Science and Technology Beijing, China, in 2021, where he is currently pursuing the Ph.D. degree with the School of Automation and Electrical Engineering. His research interests include physics-informed machine learning and neural rendering.
\end{IEEEbiography}

\vspace{-34pt}

\begin{IEEEbiography}[{\includegraphics[width=1in,height=1.25in,clip,keepaspectratio]{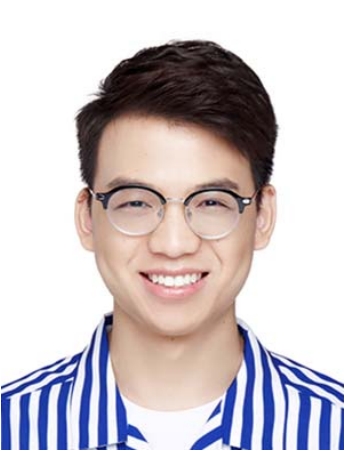}}]{Zhongwei Qiu}
is currently an algorithm expert at Alibaba DAMO Academy and a researcher at Zhejiang University. He received his PhD degree in control science and engineering from the University of Science and Technology Beijing in 2023, and his B.S. degree in automation from the University of Science and Technology Beijing in 2018. He was a visiting research scholar at the University of Sydney from 2022 to 2023. He has authored or co-authored over 10 papers in prestigious journals and top-tier conferences, such as TPAMI, CVPR, NeurIPS, ECCV, AAAI, and so on. His research is principally focused on computer vision and multimodal learning, including human-centric visual perception and generation, low-level vision and generation, as well as the application of AI for Medicine and Science.
\end{IEEEbiography}

\vspace{-34pt}


\begin{IEEEbiography}[{\includegraphics[width=1in,height=1.25in,clip,keepaspectratio]{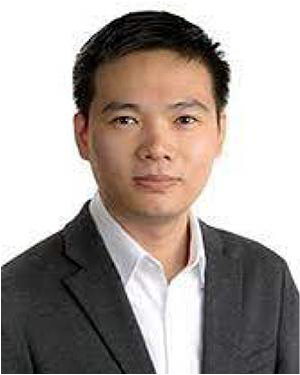}}]{Chang Xu}
received the PhD degree from Peking University, China. He is currently associate professor and ARC DECRA fellow with the School of Computer Science, University of Sydney. He has authored or coauthored more than 100 papers in prestigious journals and top tier conferences. His research interests include machine learning algorithms and related applications in computer vision. He was the recipient of several paper awards, including Distinguished Paper Award in IJCAI 2018. He was the PC member or senior PC member for many conferences, including NeurIPS, ICML, ICLR, CVPR, ICCV, IJCAI, and AAAI. He has been recognized as Top Ten Distinguished Senior PC Member in IJCAI 2017.
\end{IEEEbiography}

\vspace{-34pt}

\begin{IEEEbiography}[{\includegraphics[width=1in,height=1.25in,clip,keepaspectratio]{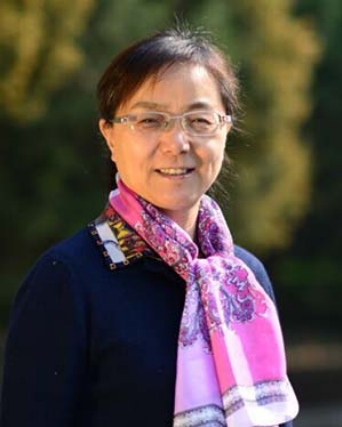}}]{Dongmei Fu}
received the M.S. degree from Northwestern Polytechnical University, in 1984, and the Ph.D. degree in automation science from the University of Science and Technology Beijing, China, in 2006, where she is currently a Professor and a Doctoral Supervisor. She has taken charge of several national projects about corrosion data mining and infrared image processing. Her current research interests include knowledge graph, image processing, and data mining.
\end{IEEEbiography}

\newpage
\clearpage

\appendices

\section{Comparison with CLIP-NeRF}

\begin{figure*}[!ht]
    \centering
    \includegraphics[width=0.95\textwidth]
    {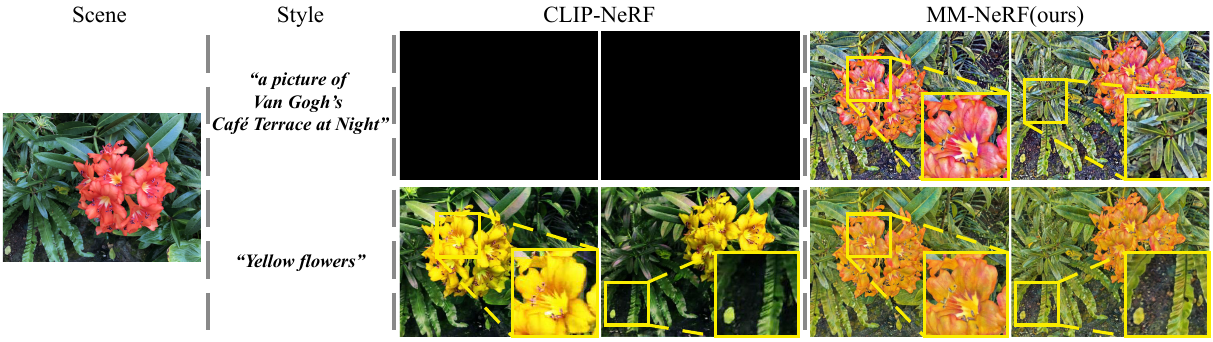}
    \vspace{-0.2cm}
    \caption{
    Comparison of text-guided 3D style transfer with CLIP-NeRF \cite{wang2022clip}. MM-NeRF transfers high-quality stylized details to 3D scenes, but CLIP-NeRF only achieves color transfer.
    }
    \vspace{-0.2cm}
    
    \label{fig:comparison_with_clip-nerf}
\end{figure*}

In Figure \ref{fig:comparison_with_clip-nerf}, we compare MM-NeRF with CLIP-NeRF \cite{wang2022clip} on forward-facing scenes \cite{mildenhall2021nerf}.
CLIP-NeRF is implemented by the released code.
In our experiments, CLIP-NeRF fails to generate stylized novel views with text: \emph{a picture of Van Gogh’s Café Terrace at Night} and CLIP loss maintains between 0.70 and 0.75. MM-NeRF can generate high-quality stylized results guided by this text.
Besides, both methods generate high-quality results guided by a photo-realistic text: \emph{Yellow flowers}.
Therefore, although CLIP-NeRF \cite{wang2022clip} supports text-guided shape and appearance editing, it only achieves color transfer based on text and cannot transfer stylized details to 3D scenes.

\section{Implementation Details}

\subsection{MM-NeRF}

\subsubsection{Neural Radiance Field}
Instant Neural Graphics Primitives (Instant-NGP) \cite{muller2022instant} is employed to implement NeRF and Ray Marching in MM-NeRF. Specifically, spatial position $\bm{x}$ and viewing direction $\bm{d}$ are encoded based on the trainable multi-resolution grid \cite{muller2022instant} and the multi-resolution sequence of spherical harmonics, respectively. The encoded spatial position is passed through two fully connected layers. Volumen density and color are predicted with two additional layers (the opacity head) and three additional layers (the color head). All layers are implemented by TCNN \cite{muller2021real} with 64 channels.

\subsubsection{Multimodal Module}
We use pre-trained CLIP (ViT-B/32) to encode styles into features.
The cross-modal feature correction module is implemented by TCNN \cite{muller2021real} and comprises six fully connected layers with 256 channels.

\subsubsection{Multi-Head Learning Scheme}
Multi-head Learning Scheme (MLS) consists of a shared backbone and independent prediction heads. The shared backbone consists of four fully connected layers, each with 512 channels, and a skip connection concatenates the input features to the third layer. MM-NeRF initializes a prediction head with the same structure for each style. In particular, the prediction head consists of two fully connected layers, each with 512 channels.

\subsection{Training}

\subsubsection{Pre-Training of Neural Radiance Field}
Adam \cite{kingma2014adam} is employed to pre-train NeRF. The learning rate begins at $2\times 10^{-2}$ and decays to $6.67\times10^{-4}$ by cosine annealing. The pre-training for each scene takes 30k iterations on a single NVIDIA GTX 4090 GPU. The batch size is 8192, 4096, and 4096 for forward-facing scenes \cite{mildenhall2021nerf}, Tanks and Temples \cite{knapitsch2017tanks}, and 360° scenes \cite{yucer2016efficient}, respectively.

\subsubsection{Pre-Training of Cross-Modal Feature Correction Module}
We use image information from WikiArt \cite{painter-by-numbers} to generate text-image pairs. 
Specifically, text is generated with the following rules: 
\begin{itemize}
    \item \emph{a picture of \{title\}},
    \item \emph{a picture of \{artist\}'s \{title\}},
    \item \emph{a picture of \{title\} by \{artist\}},
    \item \emph{a picture of \{title\} in the style of \{style\}},
    \item \emph{a picture of \{artist\}'s \{title\} in the style of \{style\}},
    \item \emph{a picture of \{title\} by \{artist\} in the style of \{style\}},
\end{itemize}
where \emph{\{title\}} is the style name, \emph{\{artist\}} is the author, \emph{\{style\}} is the style type. Missing information items are excluded. Images are generated with Stable Diffusion \cite{rombach2022high}. Specifically, stable-diffusion-v1-4 from HuggingFace \cite{von-platen-etal-2022-diffusers} is employed.
We generate 42,307 text-image pairs. The training, validation, and test sets are allocated with $0.6:0.2:0.2$. Training loss is as follows:
\begin{equation}
\small
\begin{split}
    \mathcal{L}_{CFCM} &= \mathcal{L}_{c} + \lambda\mathcal{L}_{m},\\
    \mathcal{L}_{c} &= \frac{1}{N_p}\sum_{i=1}^{N_p}(1- \frac{\bm{f}_{s_i}^T \cdot (\bm{f}_{s_t} + \mathcal{F}(\bm{f}_{s_t}))}{\Vert \bm{f}_{s_i} \Vert_2 \cdot \Vert (\bm{f}_{s_t} + \mathcal{F}(\bm{f}_{s_t})) \Vert_2}),\\
    \mathcal{L}_{m} &= \frac{1}{N_p}\sum_{i=1}^{N_p}\Vert \bm{f}_{s_i} - (\bm{f}_{s_t} + \mathcal{F}(\bm{f}_{s_t}))\Vert_2^2,
\end{split}    
\end{equation}
where $\mathcal{L}_{c}$ is the similarity loss, $\mathcal{L}_{m}$ is the mean-squared loss, $\lambda=0.5$ is the loss weight of $\mathcal{L}_{m}$, $N_p$ is the number of text-image pairs, $\mathcal{F}(.)$ is the cross-modal feature correction module, $\bm{f}_{s_i}$ is the CLIP \cite{radford2021learning} feature of a image, and $\bm{f}_{s_t}$ is the CLIP \cite{radford2021learning} feature of text. 
Adam \cite{kingma2014adam} is employed to optimize
the cross-modal feature correction module. The learning rate begins at $4\times 10^{-4}$ and decays to $1.33\times 10^{-5}$ by cosine annealing. The batch size is 256. The pre-training takes 40k iterations on a single NVIDIA GTX 4090 GPU.

\subsubsection{Pre-Training of Multi-Head Learning Scheme}
Stylization is sensitive to NeRF parameters, which means that MLS has a complex parameter space.
Minor changes in MLS predictions may also lead to drastic changes in stylized results.
Therefore, randomly initialized MLS causes extreme instability.
The above issue can be solved by reducing the learning rate, but this method decreases convergence speed.
By contrast, introducing the pre-training of MLS to ensure a better initialization is the most appropriate method to stabilize the stylization training.
Specifically, at each iteration of the pre-training of MLS, we sample a style from the style set $S$ in order and predict the parameters of the color head and the opacity head of NeRF.
MLS is optimized to minimize mean-squared loss between predicted parameters and pre-trained parameters of NeRF:
\begin{equation}
\small
    \mathcal{L}_p = \frac{1}{N_s}\sum_{i=1}^{N_s} \Vert \hat{\bm{p}}^{(i)} - \bm{p} \Vert_2^2,
    \label{eq:PreTrainingLossSPP}
\end{equation}
where $N_s$ is the total number of styles in $S$, $\hat{\bm{p}}^{(i)}$ is the predicted parameters with $s^{(i)}$ by MLS and $\bm{p}$ is the pre-trained parameters of the color head and the opacity head of NeRF.
MLS is pre-trained 400 epochs with Adam \cite{kingma2014adam} on a single NVIDIA GTX 4090 GPU. The learning rate begins at $1\times 10^{-4}$ and decays to $3.33 \times 10^{-5}$ by cosine annealing.

\subsubsection{Stylization Training}
Adam \cite{kingma2014adam} is employed to optimize MLS. The learning rate begins at $5\times 10^{-5}$ and decays to $1.67\times 10^{-6}$ by cosine annealing. Each scene is optimized with 50k iterations on a single NVIDIA GTX 4090 GPU. The batch size is 8192, 4096, and 4096 for forward-facing scenes \cite{mildenhall2021nerf}, Tanks and Temples \cite{knapitsch2017tanks}, and 360° scenes \cite{yucer2016efficient}, respectively. We pre-generate all stylized images to improve training efficiency.

\begin{figure}[t]
    \centering
    \includegraphics[width=0.99\linewidth]
    {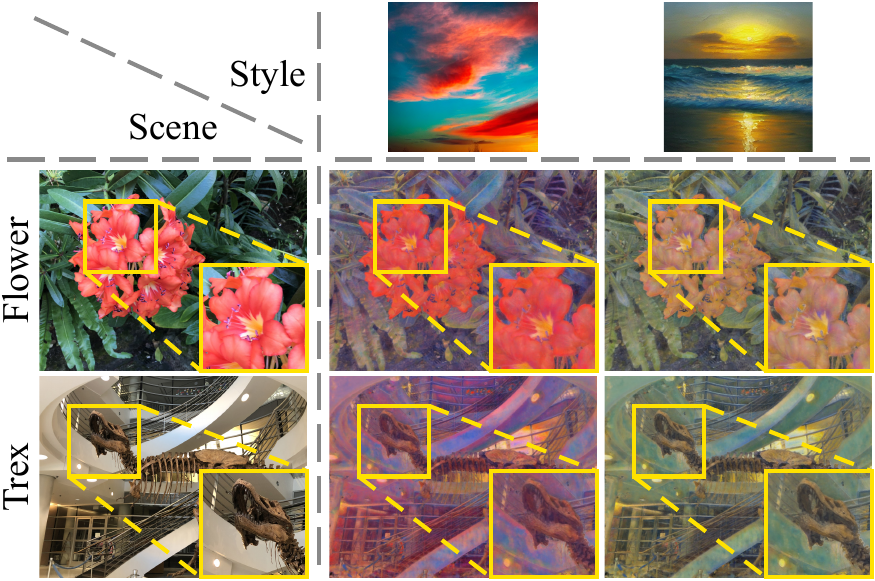}
    \vspace{-0.2cm}
    \caption{
    Failure Cases.
    MM-NeRF tends to generate blurry results with styles lacking brushstrokes.
    }
    \vspace{-0.2cm}
    
    \label{fig:failure_cases}
\end{figure}

\subsection{Multi-View Style Consistent Loss}

Masks of Multi-view Style Consistent Loss (MSCL) is computed by RAFT \cite{teed2020raft}.
RAFT consists of 3 sub-modules: 1) a feature encoder, 2) a correlation layer, and 3) an update operator.
The feature encoder is employed to extract per-pixel features from two input images.
The correlation layer computes the correlation volume of all pairs of feature vectors.
Finally, the update operator updates the predicted optical flow via GRU.
In MM-NeRF, RAFT is a fixed process. RAFT is used to compute optical flows, which are further used to compute warping function $W^{(j, ref)}(.)$ and the mask $M^{(j, ref)}$, where $j$ is the $j-th$ view and $ref$ is the reference view.

Given two images, the warping function is a bilinear interpolation, which warps the second image back to the first image according to the optical flow. 
The forward-backward consistency check is employed to compute the mask. First, we compute the optical flow $\bm{f}^{(ref, j)}$ from the $j$-th view to the reference view and the optical flow $\bm{f}^{(j, ref)}$ from the reference view to the $j$-th view. Then, the $\bm{f}^{(j, ref)}$ is further warped to the reference view via bilinear interpolation, and the result is denoted as $\tilde{\bm{f}}^{(j, ref)}$. 
We mask as occlusion where the following inequality holds:
\begin{equation}
    \Vert \bm{f}^{(ref, j)} + \tilde{\bm{f}}^{(j, ref)} \Vert_2^2 > 1.
\end{equation}
The mask $M^{(j, ref)}$ is equal to one minus the occlusion.

\begin{figure*}[!ht]
    \centering
    \includegraphics[width=0.95\textwidth]
    {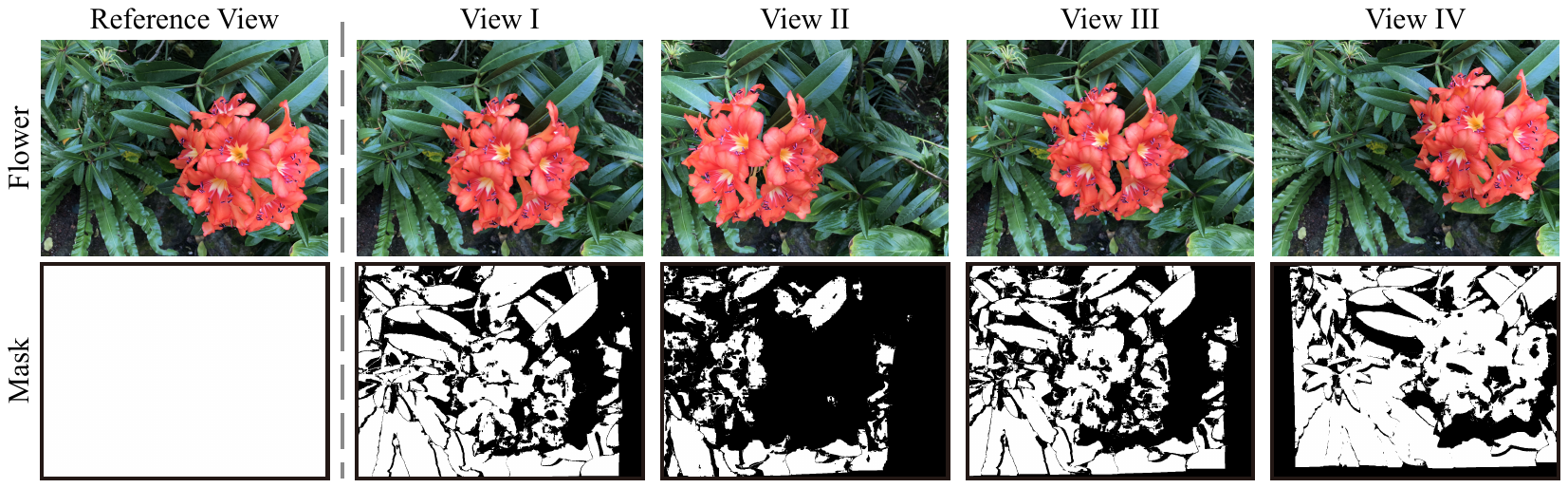}
    \caption{
    Masks of forward-facing scenes~\cite{mildenhall2021nerf}. Masks are accurately generated and used for masked reconstruction in MSCL.
    }
    
    \label{fig:flower_mask}
\end{figure*}

\begin{figure*}[!ht]
    \centering
    \includegraphics[width=0.95\textwidth]
    {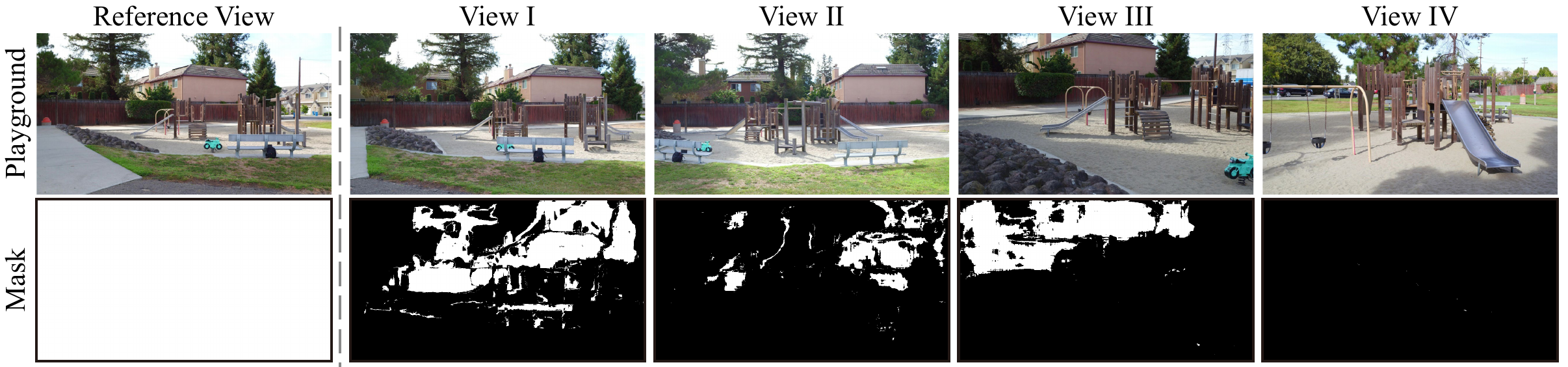}
    \caption{
    Masks of Tanks and Temples~\cite{knapitsch2017tanks}. 
    A zero mask is generated when there is a significant difference between the sampled and reference views.
    }
    
    \label{fig:tnt_mask}
\end{figure*}

Figure \ref{fig:flower_mask} shows the masks of forward-facing scenes~\cite{mildenhall2021nerf}. Due to the small difference of views, MSCL works well.
Figure \ref{fig:tnt_mask} shows the masks of Tanks and Temples~\cite{knapitsch2017tanks}. Zero masks are generated for cases with a significant difference in viewing angles between the sampled and reference views.
MSCL does not introduce error information in this case but degenerates into a common stylization loss due to zero mask.
Therefore, the MSCL always outperforms common stylization loss.
However, for large-scale scenes, as each view only covers a small part of the scene, MSCL may seriously degenerate in this case.
The stylization of large-scale scenes is a more challenging task that is out of the scope of this manuscript. We will explore it in future work.

\section{Control of Stylization}

\begin{figure}[t]
    \centering
    \includegraphics[width=0.99\linewidth]{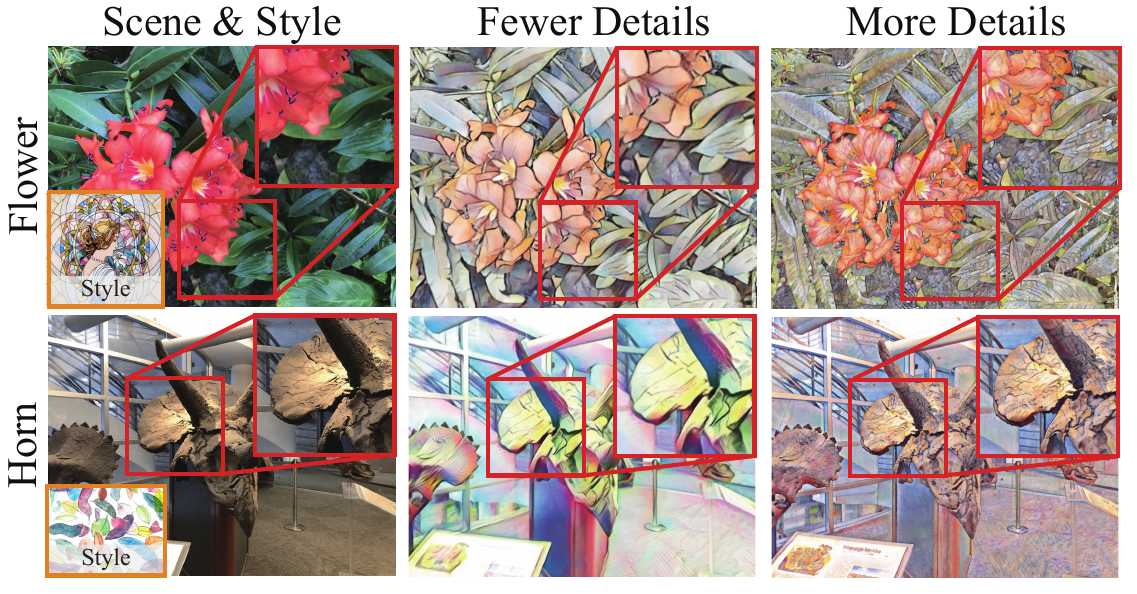}
    \caption{MM-NeRF can extend to generate stylization results with fewer details.
    }
    
    \label{fig:level_of_detail}
\end{figure}

\begin{figure*}[t]
    \centering
    \includegraphics[width=0.99\textwidth]{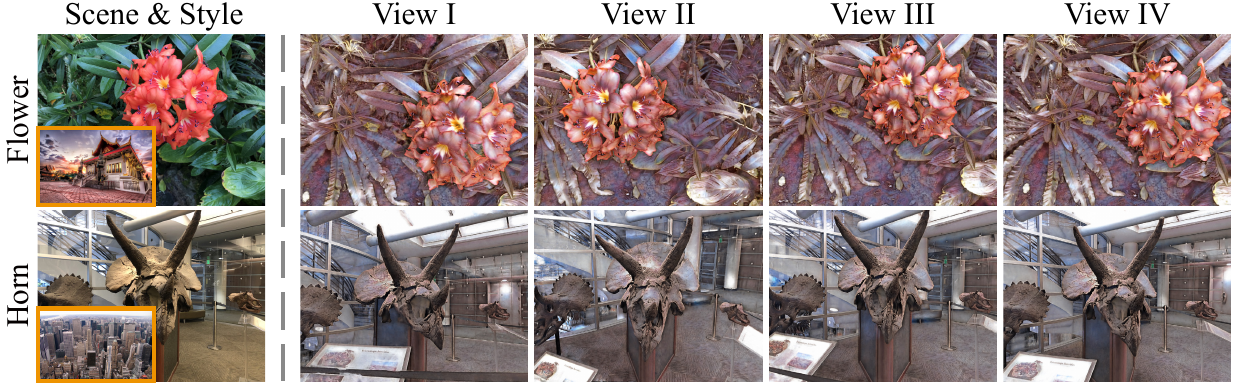}
    \caption{Photo-realistic stylization. MM-NeRF can extend to generate high-quality photo-realistic stylized novel views.
    }
    
    \label{fig:photo-realistic}
\end{figure*}

MM-NeRF employs the pre-trained AdaIN~\cite{huang2017arbitrary} to generate stylized views, which are further used to compute MSCL. Thus, the stylized details and the stylization strength of MM-NeRF can be controlled by replacing AdaIN with other methods. In our experiments, we also observed that the stylization can be controlled by scaling training views. As shown in Figure \ref{fig:level_of_detail}, MM-NeRF can extend to generate stylization results with fewer details and higher style strength than the main experiments. As shown in Figure \ref{fig:photo-realistic}, MM-NeRF can also extend to generate photo-realistic stylized novel views, which retain most of the details.
 
Depth perception and object recognition are critical for 3D scenes. In 3D stylization, we still hope to ensure the distinguishability of objects. Therefore, in our main experiments, we balance the strength and details of stylization.

\section{Calculation of Metrics}

We follow the usual practice to evaluate multi-view consistency by Temporal Warping Error (TWE) and warped Learned Perceptual Image Patch Similarity (LPIPS) \cite{huang2021learning, chiang2022hyper, nguyen2022snerf}. 
Previous methods \cite{huang2021learning, chiang2022hyper, nguyen2022snerf} render stylized videos and evaluate short-range and long-range consistency.
As MM-NeRF does not require the use of Normalized Device Coordinates (NDC), while StyleRF \cite{liu2023stylerf} asks to use it, it is difficult to render identical videos based on different methods.
To ensure the fairness of the comparison, we evaluate multi-view consistency on the test dataset.

TWE and warped LPIPS are based on optical flow, and RAFT \cite{teed2020raft} is employed to compute optical in our experiments.
The calculation of evaluation metrics is similar to our proposed MSCL. 
However, the key distinction lies in MSCL's specification of a reference view and its requirement for the consistency of details across other views with this reference view, whereas the evaluation metrics only select two distinct views and calculate the consistency of stylized results of these two views.
From the functional perspective, MSCL and evaluation metrics also exhibit significant differences. Specifically, MSCL aims to ensure the multi-view consistency of training data to prevent detail blurring and thus is utilized for processing training data, whereas evaluation metrics are employed to evaluate the multi-view consistency of generated multi-view images.

In particular, the sequence of stylized test views is generated, and TWE and warped LPIPS are computed with each pair of adjacent views. TWE is calculated as follows:
\begin{equation}
    \small
    E_{TWE} = \frac{1}{N_t-1}\sum_{i=1}^{N_t-1} \Vert I^{(i)} -  I^{\prime(i)}\Vert_2^2,
\end{equation}
where $N_t$ is the number of test views, $I^{(i)}$ is the $(i)$-th stylized view, and $I^{\prime(i)}$ is the reconstructed $(i)$-th stylized view with $(i+1)$-th stylized view. 
Similarly, warped LPIPS is computed as follows:
\begin{equation}
    \small
    E_{LPIPS} = \frac{1}{N_t-1}\sum_{i=1}^{N_t-1} LPIPS( I^{(i)}, I^{\prime(i)}),
\end{equation}
where $LPIPS(.)$ is the function to calculate LPIPS.

\section{Limitations}

Figure \ref{fig:failure_cases} shows some failure cases.
In MM-NeRF, AdaIN \cite{huang2017arbitrary} is employed to generate stylized images to optimize MM-NeRF.
When AdaIN fails, MM-NeRF cannot generate high-quality stylized results either.

Besides, the text-guided style transfer of MM-NeRF further relies on Stable Diffusion.
Therefore, when the Stable Diffusion fails to generate high-quality text-image pairs, the quality of MM-NeRF is also influenced.

\section{Additional Results}

\subsubsection{Video}
3D stylization results are best viewed as videos, and we highly recommend readers view our supplementary video for intuitive comparison.

\subsubsection{Comparison}
\begin{figure*}[t]
    \centering
    \includegraphics[width=0.99\textwidth]
    {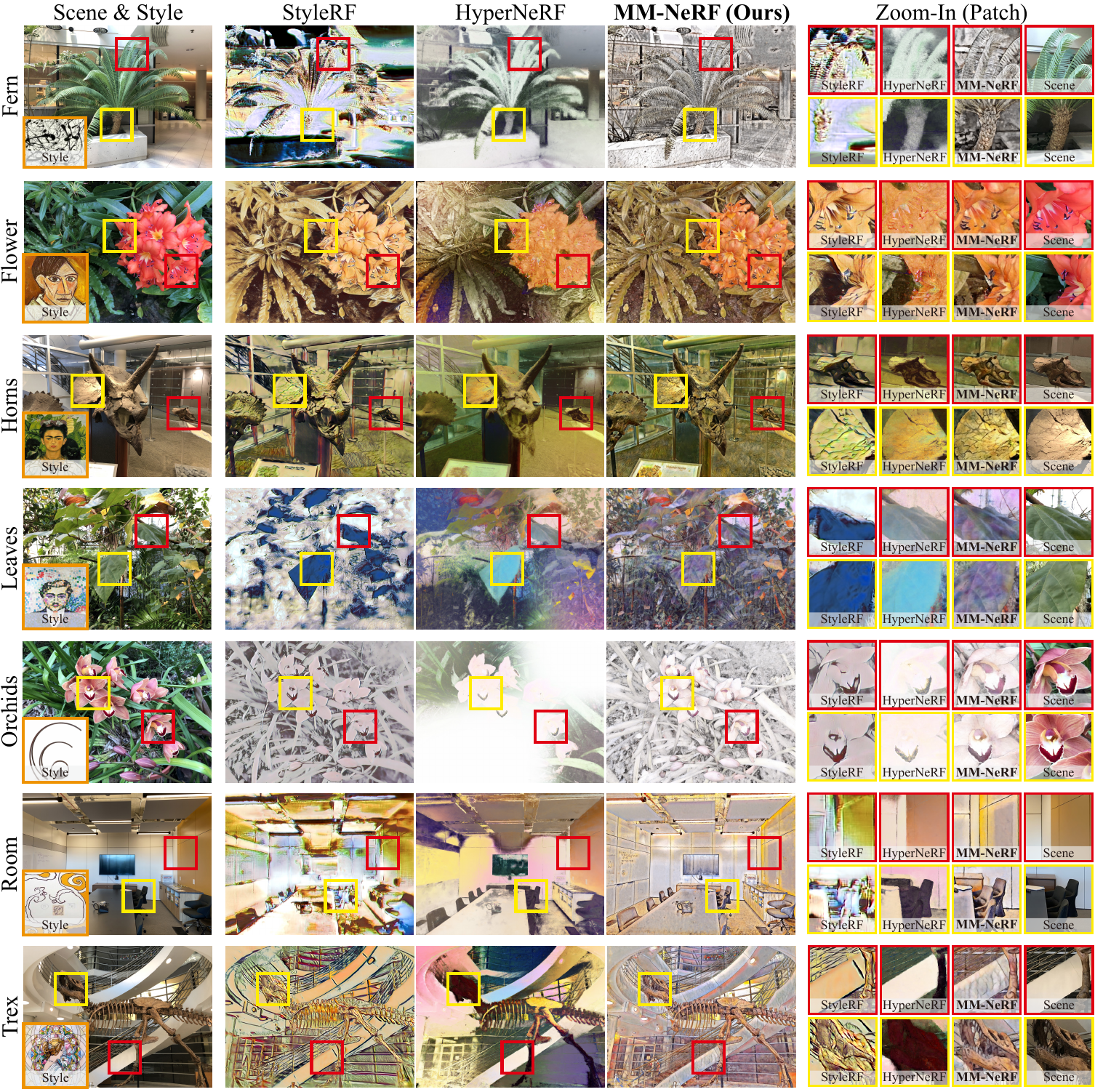}
    \caption{
    Additional Comparison with StyleRF \cite{liu2023stylerf} and HyperNeRF \cite{chiang2022hyper} on forward-facing scenes \cite{mildenhall2021nerf}.
    }
    
    \label{fig:llff}
\end{figure*}

\begin{figure*}[t]
    \centering
    \includegraphics[width=0.99\textwidth]
    {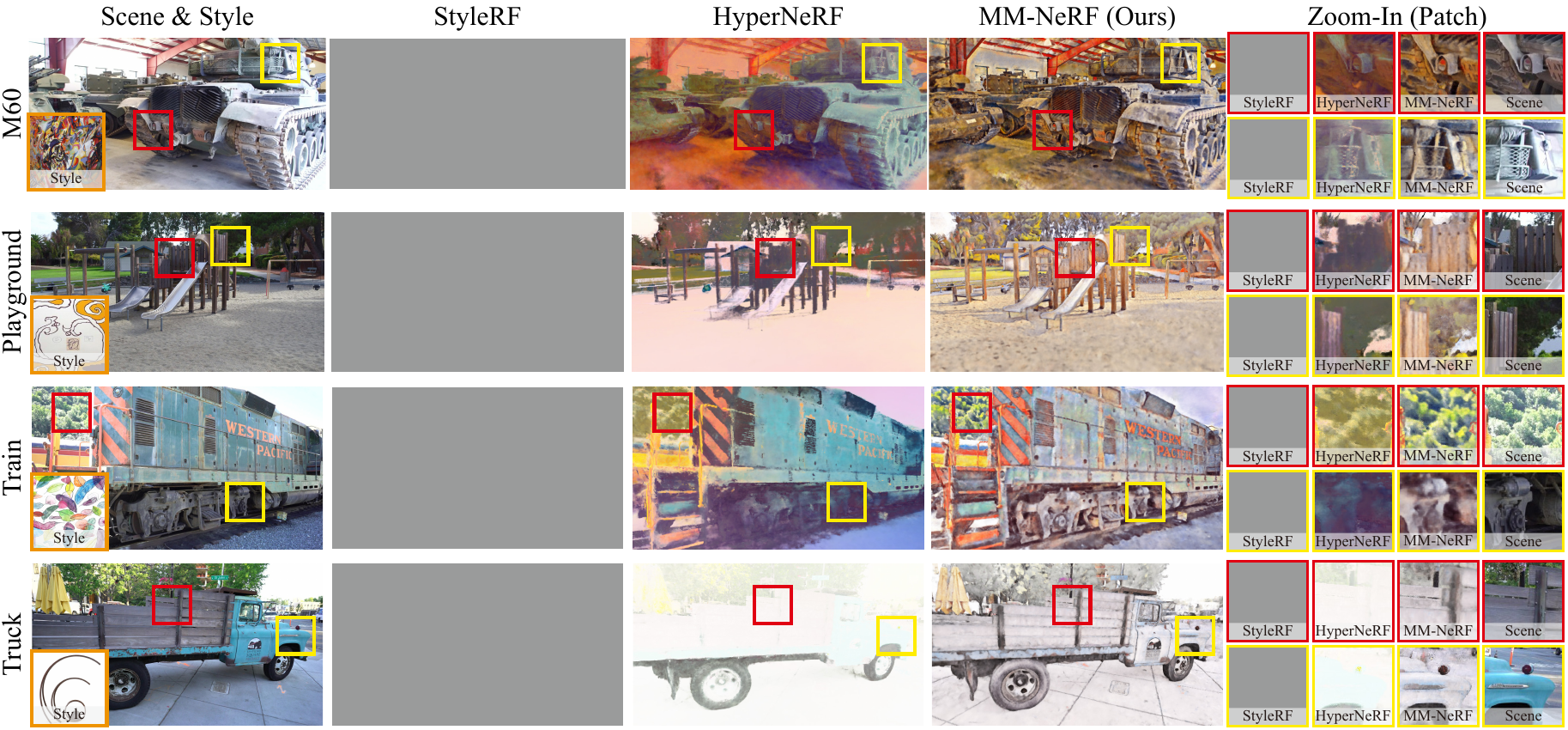}
    \caption{
    Additional Comparison with HyperNeRF \cite{chiang2022hyper} on Tanks and Temples \cite{knapitsch2017tanks}. (StyleRF cannot adapt to this dataset.)
    }
    
    \label{fig:tnt}
\end{figure*}

\begin{figure*}[t]
    \centering
    \includegraphics[width=0.99\textwidth]
    {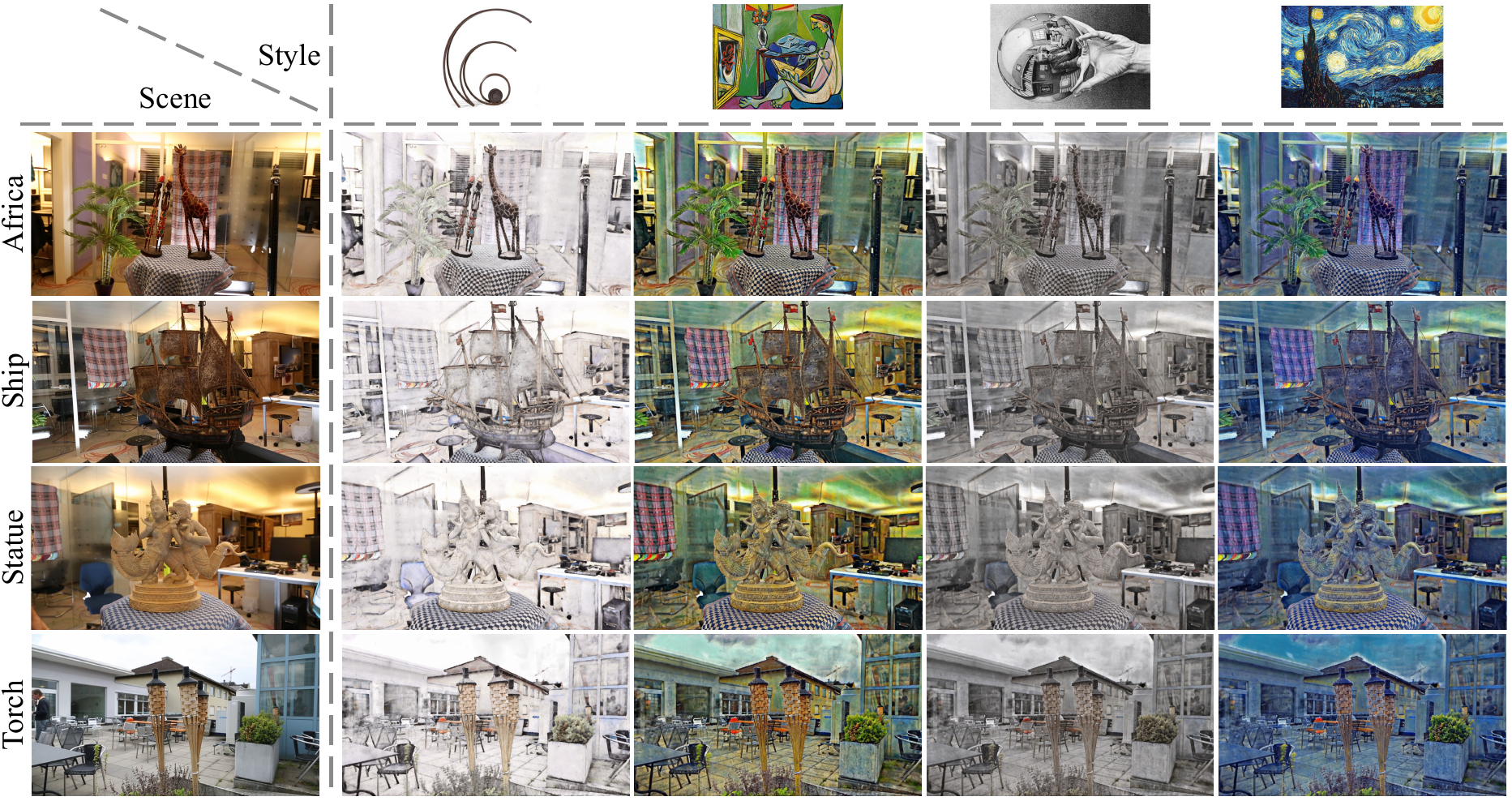}
    \caption{
    Additional results of MM-NeRF on 360° scenes \cite{yucer2016efficient}.
    }
    
    \label{fig:360}
\end{figure*}

Table \ref{Tb:multi_view_cons_llff} and Table \ref{Tb:multi_view_cons_tnt} report the multi-view consistency of each scene in the forward-facing scenes \cite{mildenhall2021nerf} and Tanks and Temples \cite{knapitsch2017tanks}, respectively.
MM-NeRF outperforms other methods on the vast majority of scenes, and the average results outperform other methods on all metrics.
Some additional results on forward-facing scenes \cite{mildenhall2021nerf} and Tank and Temples \cite{knapitsch2017tanks} are shown in Figure \ref{fig:llff} and Figure \ref{fig:tnt}, respectively.
Compared to StyleRF \cite{liu2023stylerf} and HyperNeRF \cite{chiang2022hyper}, MM-NeRF generates higher-quality stylized novel views with clearer details.
In addition, some cases on 360° scenes are shown in Figure \ref{fig:360}.
MM-NeRF also achieves high-quality stylization on this type of 3D scene.

\subsubsection{Ablation Study}
Table \ref{Tb:ablation_detail} reports the ablation study of each scene in the forward-facing scenes \cite{mildenhall2021nerf}. MLS and MSCL can effectively improve the quality of stylization on all scenes.

\clearpage
\begin{table*}[!ht]
    \centering
    \caption{
    Comparison of multi-view consistency with SOTA methods on forward-facing scenes in TWE($\downarrow$) and warped LPIPS($\downarrow$). 
    }
    \label{Tb:multi_view_cons_llff}
    \begin{tabular*}{\hsize}{@{}@{\extracolsep{\fill}}c c c c c c c c c c@{}}
    \hline

    \hline
    \multicolumn{1}{c}{\multirow{2}{*}{Method}} & \multicolumn{9}{c}{TWE$\times 10^{-2}$($\downarrow$)}\\ \cline{2-10} & fern & flower & fortress & horns & leaves & orchids & room & trex & Avg.\\
    \hline
        NeRF $\rightarrow$ AdaIN & 4.00 & 3.36 & 7.74 & 5.20 & 5.69 & 6.09 & 5.13 & 5.58 & 5.35\\
        StyTR$^2$~\cite{deng2022stytr2} & 2.91 & 3.40 & 4.78 & 4.55 & 4.64 & 4.49 & 5.19 & 4.99 & 4.37\\
        MCCNet~\cite{deng2021arbitrary} & 1.60 & 1.51 & 3.26 & 2.42 & 2.22 & 2.09 & 2.69 & 2.96 & 2.34\\
    \hline
        StyleRF \cite{chiang2022hyper} & 1.46 & 1.40 & 1.15 & 1.62 & 0.95 & 2.53 & 1.14 & 2.24 & 1.56\\
        HyperNeRF \cite{liu2023stylerf} & 1.19 & 1.81 & 1.05 & 0.95 & 0.96 & 2.49 & 1.34 & 2.55 & 1.54\\
    \hline
    
    MM-NeRF (Ours) & 1.29 & 0.87 & 1.09 & 0.88 & 1.47 & 1.70 & 1.36 & 1.27 & 1.24\\
    
    \hline
    \end{tabular*}

    \vspace{0.3cm}

    \begin{tabular*}{\hsize}{@{}@{\extracolsep{\fill}}c c c c c c c c c c@{}}
    \hline

    \hline
    \multicolumn{1}{c}{\multirow{2}{*}{Method}} & \multicolumn{9}{c}{warped LPIPS$\times 10^{-2}$($\downarrow$)}\\ \cline{2-10} & fern & flower & fortress & horns & leaves & orchids & room & trex & Avg.\\
    \hline
        NeRF $\rightarrow$ AdaIN & 8.68 & 3.95 & 9.32 & 7.11 & 13.64 & 6.13 & 8.07 & 5.33 & 7.78 \\
        StyTR$^2$~\cite{deng2022stytr2} & 8.81 & 4.10 & 8.73 & 7.17 & 14.60 & 8.06 & 8.34 & 6.39 & 8.28 \\
        MCCNet~\cite{deng2021arbitrary} & 8.39 & 4.72 & 9.62 & 7.18 & 11.97 & 7.14 & 7.81 & 6.22 & 7.88\\
    \hline
        StyleRF \cite{chiang2022hyper} & 7.98 & 4.53 & 8.18 & 6.42 & 9.84 & 6.97 & 9.01 & 5.22 & 7.27\\
        HyperNeRF \cite{liu2023stylerf} & 9.10 & 5.04 & 10.29 & 7.18 & 11.39 & 8.45 & 8.20 & 6.74 & 8.30\\
    \hline
    MM-NeRF (Ours) & 7.87 & 4.28 & 11.08 & 5.45 & 10.71 & 5.72 & 6.36 & 4.92 & 7.05\\
    
    \hline
    \end{tabular*}
\end{table*}

\begin{table*}[!ht]
    \centering
    \caption{
    Comparison of multi-view consistency with SOTA methods on Tanks and Temples in TWE($\downarrow$) and warped LPIPS($\downarrow$). (StyleRF cannot adapt to this dataset.)
    }
    \label{Tb:multi_view_cons_tnt}
    \begin{tabular*}{\hsize}{@{}@{\extracolsep{\fill}}c c c c c c@{}}
    \hline

    \hline
    \multicolumn{1}{c}{\multirow{2}{*}{Method}} & \multicolumn{5}{c}{TWE$\times 10^{-2}$($\downarrow$)}\\ \cline{2-6} & M60 & Playground & Train & Truck & Avg.\\
    \hline
        NeRF $\rightarrow$ AdaIN & 3.91 & 3.72 & 5.84 & 4.16 & 4.41\\
        StyTR$^2$~\cite{deng2022stytr2} & 1.91 & 1.74 & 2.09 & 1.76 & 1.88 \\
        MCCNet~\cite{deng2021arbitrary} & 2.50 & 2.11 & 3.13 & 2.44 & 2.54\\
    \hline
        StyleRF \cite{liu2023stylerf} & - & - & - & - & -\\
        HyperNeRF \cite{chiang2022hyper} & 0.42 & 0.49 & 0.86 & 0.49 & 0.57\\
    \hline
    MM-NeRF (Ours) & 0.42 & 0.35 & 0.75 & 0.45 & 0.49\\
    
    \hline
    \end{tabular*}

    \vspace{0.3cm}

    \begin{tabular*}{\hsize}{@{}@{\extracolsep{\fill}}c c c c c c@{}}
    \hline

    \hline
    \multicolumn{1}{c}{\multirow{2}{*}{Method}} & \multicolumn{5}{c}{warped LPIPS$\times 10^{-2}$($\downarrow$)}\\ \cline{2-6} & M60 & Playground & Train & Truck & Avg.\\
    \hline
        NeRF $\rightarrow$ AdaIN & 13.17 & 14.07 & 12.18 & 11.08 & 12.63\\
        StyTR$^2$~\cite{deng2022stytr2} & 14.50 & 14.38 & 14.79 & 9.64 & 13.33 \\
        MCCNet~\cite{deng2021arbitrary} & 10.84 & 11.02 & 10.89 & 9.75 & 10.63 \\
    \hline
        StyleRF \cite{liu2023stylerf} & - & - & - & - & -\\
        HyperNeRF \cite{chiang2022hyper} & 9.82 & 7.52 & 8.72 & 9.83 & 8.97\\
    \hline
    MM-NeRF (Ours) & 10.11 & 8.22 & 8.34 & 8.73 & 8.85\\
    
    \hline
    \end{tabular*}
\end{table*}

\begin{table*}[!ht]
    \centering
    \caption{
    Ablation study of MSCL and MSL on forward-facing scenes with LPIPS($\downarrow$), PSNR($\uparrow$), and SSIM($\uparrow$). 
    }
    \label{Tb:ablation_detail}
    \begin{tabular*}{\hsize}{@{}@{\extracolsep{\fill}}c c c c c c c c c c@{}}
    \hline

    \hline
    \multicolumn{1}{c}{\multirow{2}{*}{Method}} & \multicolumn{9}{c}{LPIPS$\times 10^{-2}$($\downarrow$)}\\ \cline{2-10} & fern & flower & fortress & horns & leaves & orchids & room & trex & Avg.\\
    \hline
    NeRF & 32.98 & 35.22 & 37.40 & 33.57 & 38.08 & 37.07 & 36.11 & 32.88 & 35.42\\
    \hline
    NeRF + MSCL & 33.14 & 35.71 & 37.26 & 33.45 & 38.63 & 37.36 & 35.96 & 33.22 & 35.59\\
    NeRF + MLS & 28.12 & 26.95 & 30.86 & 27.36 & 29.48 & 31.85 & 29.66 & 27.74 & 29.00\\
    MM-NeRF & 27.85 & 26.20 & 29.08 & 26.02 & 29.20 & 31.95 & 29.15 & 27.17 & 28.33\\
    
    \hline
    \end{tabular*}

    \vspace{0.3cm}

    \begin{tabular*}{\hsize}{@{}@{\extracolsep{\fill}}c c c c c c c c c c@{}}
    \hline

    \hline
    \multicolumn{1}{c}{\multirow{2}{*}{Method}} & \multicolumn{9}{c}{PSNR($\uparrow$)}\\ \cline{2-10} & fern & flower & fortress & horns & leaves & orchids & room & trex & Avg.\\
    \hline
    NeRF & 19.40 & 19.76 & 19.21 & 19.88 & 19.32 & 17.92 & 19.90 & 19.67 & 19.38\\
    \hline
    NeRF + MSCL & 19.67 & 20.00 & 19.54 & 20.16 & 19.60 & 17.97 & 19.99 & 19.84 & 19.60\\
    NeRF + MLS & 19.27 & 20.03 & 19.39 & 20.51 & 19.50 & 17.77 & 20.48 & 20.11 & 19.63\\
    MM-NeRF & 19.29 & 20.46 & 20.00 & 21.02 & 19.94 & 17.58 & 20.70 & 20.41 & 19.92\\
    
    \hline
    \end{tabular*}

    \vspace{0.3cm}

    \begin{tabular*}{\hsize}{@{}@{\extracolsep{\fill}}c c c c c c c c c c@{}}
    \hline

    \hline
    \multicolumn{1}{c}{\multirow{2}{*}{Method}} & \multicolumn{9}{c}{SSIM$\times 10^{-2}$($\uparrow$)}\\ \cline{2-10} & fern & flower & fortress & horns & leaves & orchids & room & trex & Avg.\\
    \hline
    NeRF & 50.65 & 50.13 & 46.56 & 52.95 & 46.32 & 41.24 & 49.61 & 55.68 & 49.14\\
    \hline
    NeRF + MSCL & 52.27 & 51.61 & 48.44 & 55.07 & 48.29 & 41.63 & 50.27 & 56.66 & 50.53\\
    NeRF + MLS & 50.28 & 53.30 & 49.38 & 56.74 & 48.35 & 41.93 & 51.51& 57.98 & 51.19\\
    MM-NeRF & 52.76 & 56.50 & 53.81 & 60.56 & 51.53 & 42.35 & 53.18 & 59.96 & 53.83\\
    
    \hline
    \end{tabular*}
\end{table*}

\clearpage

\vfill

\end{document}